\documentclass[table]{article}

\usepackage[main, final]{neurips_2026}

\usepackage{amsmath,amsfonts,bm}

\def\eqref#1{equation~\ref{#1}}

\def\rvp{{\mathbf{p}}}

\def\rvw{{\mathbf{w}}}

\def\ervw{{\textnormal{w}}}

\DeclareMathAlphabet{\mathsfit}{\encodingdefault}{\sfdefault}{m}{sl}
\SetMathAlphabet{\mathsfit}{bold}{\encodingdefault}{\sfdefault}{bx}{n}

\newcommand{\softmax}{\mathrm{softmax}}

\usepackage[utf8]{inputenc} 
\usepackage[T1]{fontenc}    
\usepackage{url}            
\usepackage{booktabs}       
\usepackage{amsfonts}       
\usepackage{nicefrac}       
\usepackage{microtype}      
\usepackage{xcolor}         

\usepackage{amsmath}
\usepackage{amssymb}
\usepackage{amsthm}
\usepackage{cases}
\usepackage{graphicx}
\usepackage[colorlinks=true, allcolors=blue]{hyperref}
\usepackage{multirow}
\usepackage[ruled,vlined,noend]{algorithm2e}
\usepackage{pifont}
\usepackage{tabularx} 
\usepackage{makecell} 
\usepackage{caption} 
\usepackage{subcaption} 
\usepackage{dsfont} 
\usepackage{xspace} 

\usepackage{styles/formatting}

\title{Jacobian Descent for Multi-Objective Optimization}

\author{%
  Pierre Quinton $^\ast$ \\
  LTHI, EPFL\\
  \texttt{pierre.quinton@epfl.ch} \\
  \And
  Valérian Rey $^\ast$ \\
  SimplexLab \\
  \texttt{valerian.rey@gmail.com} \\
}

\begin{document}

\maketitle

\def\thefootnote{*}\footnotetext{Equal contribution}\def\thefootnote{\arabic{footnote}}

\bibliographystyle{styles/references}

\renewcommand\qedsymbol{}

\begin{abstract}
Many optimization problems require balancing multiple conflicting objectives.
To generalize gradient descent, which is limited to single-objective optimization, we introduce Jacobian descent (JD) and its stochastic variants.\footnote{Our library enabling JD with \pytorch\ is available at \link{https://github.com/SimplexLab/TorchJD}}
This algorithm iteratively updates parameters using the Jacobian matrix of a vector-valued objective function, in which each row is the gradient of an individual objective.
While several methods to combine gradients already exist in the literature, they generally struggle when the objectives conflict.
In contrast, we propose projecting gradients to fully resolve conflict.
We prove convergence to the Pareto front in the smooth convex case with this approach, a stronger guarantee than the usual convergence to a weakly Pareto stationary point.
Our method also enables instance-wise risk minimization (IWRM), a novel learning paradigm in which the loss of each training example is considered a separate objective.
Early experimentation on image classification shows promising results for IWRM.
Lastly, we provide an efficient implementation of JD using the Gramian of the Jacobian matrix to drastically reduce memory requirements.
\end{abstract}
{
\colorbox{green!3}{\begin{minipage}{\dimexpr\textwidth-2\fboxsep\relax}
\begin{minipage}{0.36\linewidth}
\begin{algorithm}[H]
\label{alg:GD}
\caption{Gradient descent$\vphantom{\A}$}
\DontPrintSemicolon
\KwIn{$\vx\in\Rn$, $0<\eta$, $T\in\mathbb{N}$$\vphantom{\A\Rmn}$}
\For{$t \gets 1$ \KwTo $T$}{
    $\vx \gets \vx - \eta \nabla f(\vx)$$\vphantom{\A\big(\jac \vf(\vx)\big)}$\;
}
\KwOut{$\vx$}
\end{algorithm}
\end{minipage}
\hfill
\begin{minipage}{0.545\linewidth}
\begin{algorithm}[H]
\label{alg:JD}
\caption{Jacobian descent with aggregator $\A$}
\DontPrintSemicolon
\KwIn{$\vx\in\Rn$, $0<\eta$, $T\in\mathbb N$, $\A:\Rmn\to\Rn$}
\For{$t \gets 1$ \KwTo $T$}{
    $\vx\gets \vx - \eta \A\big(\jac \vf(\vx)\big)$\;
}
\KwOut{$\vx$}
\end{algorithm}
\end{minipage}
\end{minipage}}%
}

\section{Introduction}
The field of multi-objective optimization studies minimization of vector-valued objective functions~\citep{moo_book_sawaragi, moo_book_ehrgott, moo_book_branke, moo_book_deb}.
In deep learning, a widespread approach to train a model with multiple objectives is to combine those into a scalar loss function minimized by stochastic gradient descent.
While this method is simple, it comes at the expense of potentially degrading some individual objectives.
Without prior knowledge of their relative importance, this is undesirable.
In contrast, multi-objective optimization methods typically attempt to optimize all objectives simultaneously, without making arbitrary compromises: the goal is to find points for which no improvement can be made on some objectives without degrading others.
Such points are called \emph{Pareto optimal}, and their image by the objective function is called the \emph{Pareto front}.

Early works have attempted to extend gradient descent (GD) to consider several objectives simultaneously, and thus several gradients~\citep{fs, mgda1}.
Essentially, they propose a heuristic to prevent the degradation of any individual objective.
Several other works have built upon this method, analyzing its convergence properties or extending it to a stochastic setting~\citep{smsmgda, smgda, fs_complexity}.
Later, this has been applied to multi-task learning to tackle conflict between tasks, illustrated by contradicting gradient directions~\citep{mgda2}.
Many studies have followed, proposing various other algorithms for the training of multi-task models~\citep{pcgrad, graddrop, cagrad, imtl, random, nashmtl, aligned_mtl}.
They commonly rely on an \emph{aggregator}: a function mapping the Jacobian matrix, whose rows are the gradients of the individual objectives, to a single parameter update.

We propose to unify all such methods under the \emph{Jacobian descent} (JD) algorithm, specified by an aggregator.
This algorithm aims to minimize a differentiable vector-valued function $\vf: \Rn \to \Rm$ iteratively without relying on a scalarization of the objective.
Intuitively, at each iteration, JD computes the gradient of each objective, combines these gradients into a single update direction using the aggregator, and takes a small step in that direction, just like GD does with a single gradient.
The role of the aggregator is thus to find a direction that is good for all objectives at once.
Under this formulation, the existing methods are simply distinguished by their aggregator.
Consequently, studying its properties is essential for understanding the behavior and convergence of JD.
Under significant conflict, existing aggregators often fail to provide strong convergence guarantees.
To address this, we propose $\upgrad$, specifically designed to resolve conflicts while naturally preserving the relative influence of individual gradients.

Furthermore, we introduce a novel stochastic variant of JD that enables the training of neural networks with a large number of objectives.
This unlocks a particularly interesting perspective: considering the minimization of instance-wise loss vectors rather than the usual minimization of the average training loss.
As this paradigm is a direct generalization of the well-known empirical risk minimization (ERM)~\citep{erm}, we name it \emph{instance-wise risk minimization} (IWRM).

\paragraph{Contributions.}

(1) We formalize the JD algorithm and its stochastic variants.
(2) We define a new aggregator $\upgrad$, based on gradient projections, and satisfying important properties.
(3) We show convergence of JD with $\upgrad$ to the Pareto front in the smooth convex case, a stronger guarantee than the one usually provided for existing methods.
(4) We define formally the IWRM learning paradigm, backed by some positive preliminary experiments.
(5) We provide a novel formulation of Jacobian descent, relying on the Gramian of the Jacobian, decreasing memory usage from $\bigo(m \times n)$ to $\bigo(n + m^2)$.
(6) We provide an open source $\pytorch$-based library to do Jacobian descent with any aggregator, with either the naive or the Gramian-based implementation.

\section{Conceptual and theoretical foundations}\label{sec:theoretical}
To optimize a vector-valued objective, a partial order between vectors must be considered.
Throughout this paper, $\pleq$ denotes the relation defined for any pair of vectors $\vu, \vv\in \Rm$ as $\vu\pleq \vv$ whenever $u_i\leq v_i$ for all coordinates $i$.
Similarly, $\plt$ is the relation defined by $\vu\plt \vv$ whenever $u_i<v_i$ for all coordinates $i$.
Furthermore, $\vu\plneq \vv$ indicates that both $\vu\pleq \vv$ and $\vu\neq \vv$ hold.
The Euclidean vector norm and the Frobenius matrix norm are denoted by $\| \cdot \|$ and $\| \cdot \|_\Frob$, respectively.
Finally, for any $m\in\mathbb N$, the symbol $\range{m}$ represents the range $\{ i\in \mathbb N : 1\leq i\leq m \}$.

\subsection{Jacobian descent}\label{subsection:jd}

In the following, we introduce Jacobian descent, a natural extension of gradient descent supporting the optimization of vector-valued functions.

Suppose that $\vf: \Rn \to \Rm$ is continuously differentiable.
Let $\jac \vf(\vx) \in \Rmn$ be the Jacobian matrix of $\vf$ at $\vx$, \ie
\begin{equation}
\jac \vf(\vx) = \begin{bmatrix}\nabla f_1(\vx)^\T\\\nabla f_2(\vx)^\T\\\vdots\\\nabla f_m(\vx)^\T\end{bmatrix} = 
\begin{bmatrix}
\frac{\partial}{\partial x_1} f_1(\vx)&\frac{\partial}{\partial x_2} f_1(\vx)&\cdots&\frac{\partial}{\partial x_n} f_1(\vx)\\
\frac{\partial}{\partial x_1} f_2(\vx)&\frac{\partial}{\partial x_2} f_2(\vx)&\cdots&\frac{\partial}{\partial x_n} f_2(\vx)\\
\vdots&\vdots&\ddots&\vdots\\
\frac{\partial}{\partial x_1} f_m(\vx)&\frac{\partial}{\partial x_2} f_m(\vx)&\cdots&\frac{\partial}{\partial x_n} f_m(\vx)\\
\end{bmatrix}
\end{equation}

Given $\vx, \vy\in\Rn$, Taylor's theorem yields
\begin{equation}
\vf(\vx+\vy) = \vf(\vx) + \jac \vf(\vx)\cdot \vy + o(\|\vy\|),\label{eqn:taylor-approximation}
\end{equation}
The term $\vf(\vx) + \jac \vf(\vx) \cdot \vy$ is the first-order Taylor approximation of $\vf(\vx+\vy)$.
Via this approximation, we aim to select a small update $\vy$ that reduces $\vf(\vx+\vy)$, ideally achieving $\vf(\vx+\vy)\pleq\vf(\vx)$.
As the approximation depends on $\vy$ only through $\jac \vf(\vx) \cdot \vy$, selecting the update based on the Jacobian is natural.
A mapping $\A:\Rmn\to\Rn$ reducing such a matrix into a vector is called an \emph{aggregator}.
For any $\J \in \Rmn$, $\A(\J)$ is called the \emph{aggregation} of $\J$ by $\A$.

To minimize $\vf$, consider the update $\vy = - \eta \A\big(\jac \vf(\vx)\big)$, where $\eta$ is an appropriate step size, and $\A$ is an appropriate aggregator.
Jacobian descent simply consists in applying this update iteratively, as shown in Algorithm~\ref{alg:JD}.
To put it into perspective, we also provide a minimal version of GD in Algorithm~\ref{alg:GD}.
Remarkably, when $m=1$, the Jacobian has a single row, so GD is a special case of JD where the aggregator is the identity.

Note that other gradient-based optimization algorithms, \eg Adam~\citep{adam}, can similarly be extended to the multi-objective case.

\paragraph{Stochastic variants.} In some settings, the exact computation of the update can be prohibitively slow or even intractable.
When dealing with a single objective, stochastic gradient descent (SGD) replaces the gradient $\nabla f(\vx)$ with some estimation.
More generally, \emph{stochastic Jacobian descent} (SJD) relies on estimates of the aggregation of the Jacobian.
One approach, that we call \emph{stochastically estimated Jacobian descent} (SEJD), is to compute and aggregate an estimation of the Jacobian.
Alternatively, when the number of objectives is very large, we propose to aggregate a matrix whose rows are a random subset of the rows of the true Jacobian.
We call this approach \emph{stochastic sub-Jacobian descent} (SSJD).

\subsection{Desirable properties for aggregators}\label{subsection:properties}

An inherent challenge of multi-objective optimization is to manage conflicting objectives~\citep{mgda2, pcgrad, cagrad}.
Substituting the update $\vy = -\eta \A\big(\jac \vf(\vx)\big)$ into the first-order Taylor approximation $\vf(\vx)~+~\jac \vf(\vx) \cdot \vy$ yields $\vf(\vx) - \eta \jac \vf(\vx) \cdot \A\big(\jac \vf(\vx)\big)$.
In particular, if $\0\pleq\jac \vf(\vx) \cdot \A\big(\jac \vf(\vx)\big)$, then no coordinate of the approximation of $\vf$ will increase.
A pair of vectors $\vx, \vy\in\Rn$ is said to \emph{conflict} if $\vx^\T \vy < 0$.
Hence, for a sufficiently small $\eta$, if any row of $\jac \vf(\vx)$ conflicts with $\A\big(\jac \vf(\vx)\big)$, the corresponding coordinate of $\vf$ will increase.
When minimizing $\vf$, avoiding conflict between the aggregation and any gradient is thus desirable, motivating the first property.

\begin{definition}[Non-conflicting]\label{def:non-conflicting}
Let $\A:\Rmn\to\Rn$ be an aggregator.
If for all $\J\in\Rmn$, $\0\pleq \J\cdot \A(\J)$, then $\A$ is said to be \emph{non-conflicting}.
\end{definition}

With a non-conflicting aggregator and a sufficiently small step size, no objective increases at any iteration of JD, even when the objectives strongly conflict.
Without this property, some objectives can be improved at the expense of others, which is precisely what multi-objective optimization aims to avoid.

For any collection of vectors $C \subseteq \Rn$, the \emph{dual cone} of $C$ is $\{ \vx\in\Rn: \forall \vy\in C, 0\leq \vx^\T \vy \}$~\citep{convex_optimization}.
Notice that an aggregator $\A$ is non-conflicting if and only if for any $\J$, $\A(\J)$ is in the dual cone of the rows of $\J$, as illustrated in Figure~\ref{fig:aggregator_directions}.

In a step of GD, the update scales proportionally to the gradient norm.
Small gradients thus lead to small updates, and conversely, large gradients lead to large updates.
To maintain coherence with GD, it would be natural that the rows of the Jacobian also contribute to the aggregation proportionally to their norm.
Scaling each row of $\jac \vf(\vx)$ by the corresponding element of some vector $\vc \in \Rm$ yields $\diag(\vc) \cdot \jac \vf(\vx)$, where $\diag(\vc)\in\mathbb R^{m\times m}$ is the diagonal matrix whose diagonal is $\vc$.
This insight can then be formalized as the following property.

\begin{definition}[Linear under scaling]\label{def:scaling_linear}
Let $\A:\Rmn \to \Rn$ be an aggregator.
If for all $\J\in\Rmn$, the mapping from any $\0\plt \vc\in\Rm$ to $\A\big(\diag(\vc)\cdot \J\big)$ is linear in $\vc$, then $\A$ is said to be \emph{linear under scaling}.
\end{definition}

Each gradient then contributes to the aggregation proportionally to its norm, so small gradients can neither dominate the update nor make it vanish.
For instance, $\mgda$, presented in Section~\ref{sec:related}, lacks this property: if any row of the Jacobian approaches zero, so does its aggregation.

Finally, as $\| \vy \|$ decreases asymptotically to $0$, the precision of the first-order Taylor approximation $\vf(\vx) + \jac \vf(\vx) \cdot \vy$ improves, as highlighted in (\ref{eqn:taylor-approximation}).
The projection $\vy'$ of any candidate update $\vy$ onto the span of the rows of $\jac \vf(\vx)$ satisfies $\jac \vf(\vx) \cdot \vy' = \jac \vf(\vx) \cdot \vy$ and $\|\vy'\| \leq \|\vy\|$ (see Lemma~\ref{lem:row_span_projection} in Appendix~\ref{appendix:preliminary}), so this projection decreases the norm of the update while preserving the value of the approximation.
Without additional information about $\vf$, it is thus reasonable to select $\vy$ directly in the row span of $\jac \vf(\vx)$, \ie to have a vector of weights $\vw \in \Rm$ satisfying $\vy = \jac\vf(\vx)^\T \cdot \vw$.
This yields the last desirable property.

\begin{definition}[Weighted]\label{def:in_span}
Let $\A:\Rmn \to \Rn$ be an aggregator.
If for all $\J\in\Rmn$, there exists $\vw \in \Rm$ satisfying $\A(\J) = \J^\T \cdot \vw$, then $\A$ is said to be \emph{weighted}.
\end{definition}

A weighted aggregator thus never wastes part of the update on directions along which, to first order, all objectives are constant.
Moreover, its aggregation is entirely described by the weights $\vw$, which give the contribution of each gradient to the update.

\subsection{Unconflicting projection of gradients}\label{subsection:upgrad}
\begin{figure}[ht]
\centering
\begin{subfigure}[b]{0.49\textwidth}
\centering
\includegraphics[width=1.0\textwidth]{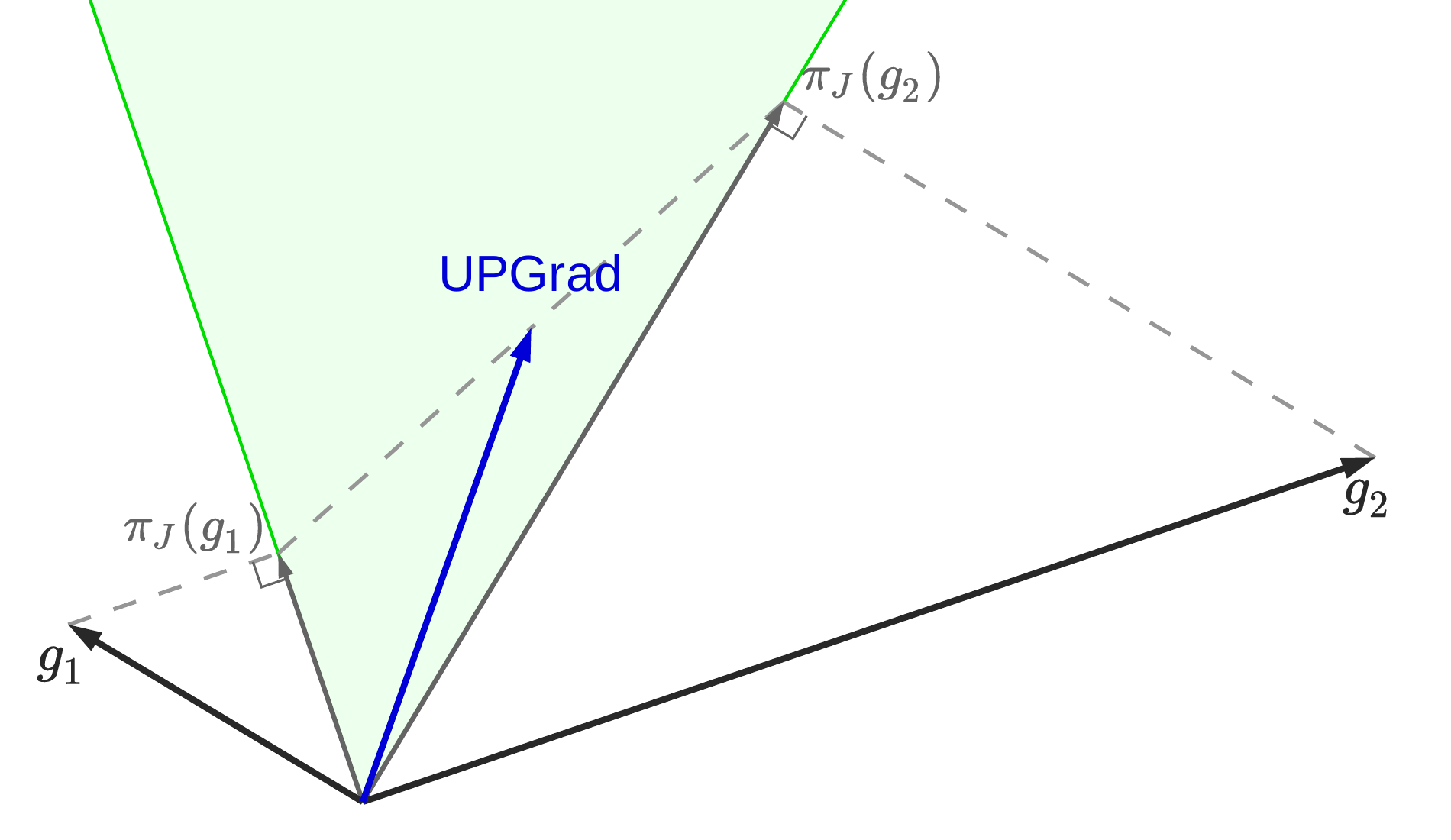}
\caption{$\upgrad(\J)$ (ours) \label{fig:upgrad_direction}}
\end{subfigure}
\begin{subfigure}[b]{0.49\textwidth}
\centering
\includegraphics[width=1.0\textwidth]{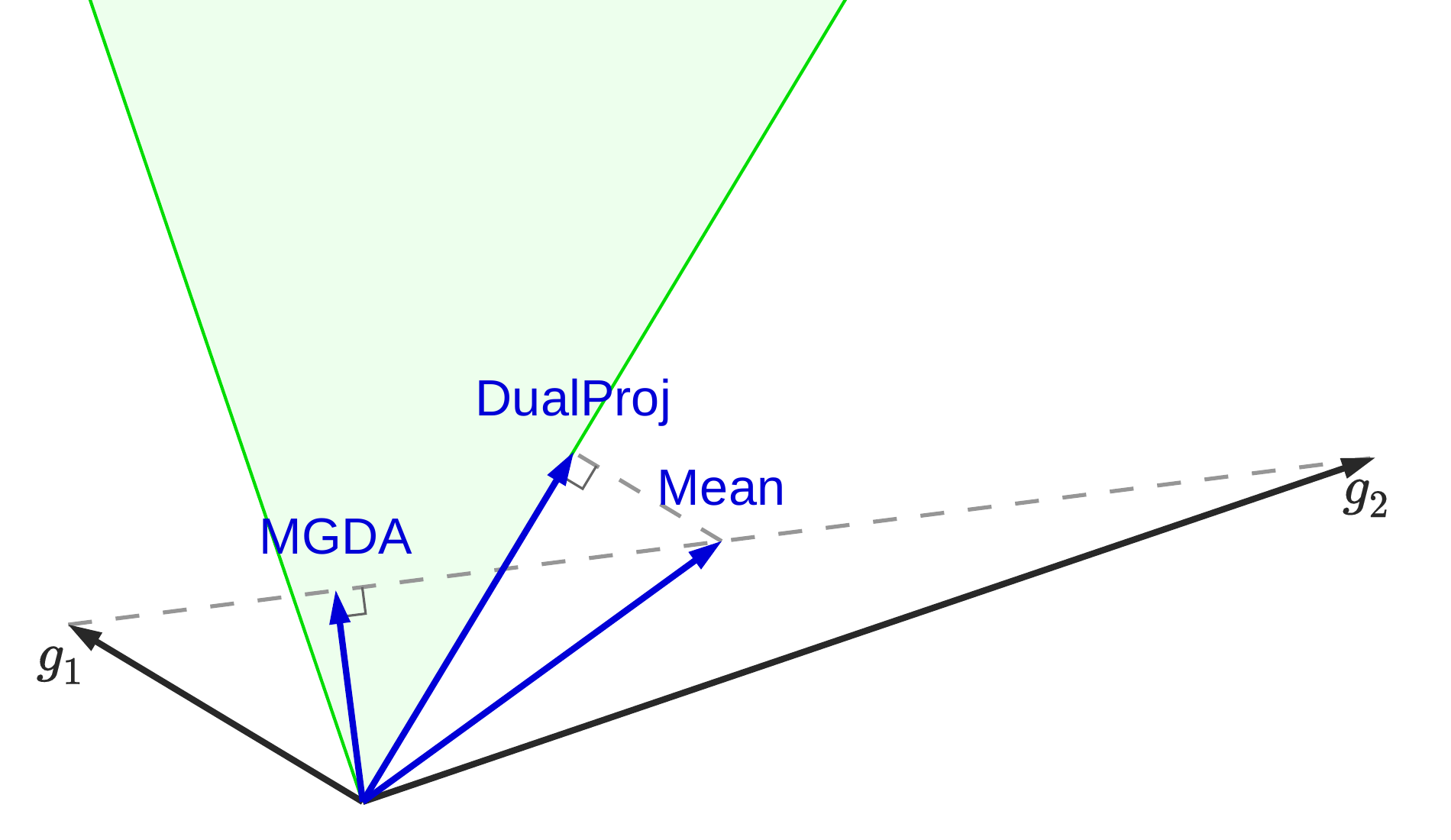}
\caption{$\meanagg(\J)$, $\mgda(\J)$ and $\dualproj(\J)$\label{fig:other_directions}}
\end{subfigure}
\caption{Aggregation of $\J=[\vg_1 \; \vg_2]^\T\in\mathbb R^{2\times 2}$ by four different aggregators. The dual cone of $\{\vg_1, \vg_2\}$ is represented in green.\\
(a) $\upgrad$ projects $\vg_1$ and $\vg_2$ onto the dual cone and averages the results.\\
(b) The mean $\meanagg(\J) = \frac{1}{2}(\vg_1 + \vg_2)$ conflicts with $\vg_1$. $\dualproj$ projects this mean onto the dual cone, so it lies on its boundary. $\mgda(\J)$ is almost orthogonal to $\vg_2$ because of its larger norm.}\label{fig:aggregator_directions}
\end{figure}

We now define the \emph{unconflicting projection of gradients} aggregator $\upgrad$, specifically designed to be non-conflicting, linear under scaling, and weighted.
In essence, it projects each gradient onto the dual cone of the rows of the Jacobian and averages the results, as illustrated in Figure~\ref{fig:upgrad_direction}.

For any $\J\in\Rmn$ and $\vx\in\Rn$, the \emph{projection} of $\vx$ onto the dual cone of the rows of $\J$ is
\begin{equation}
\proj_\J(\vx)= \argmin_{\substack{\vy\in\Rn:~\0\pleq \J \vy}} \| \vy-\vx \|^2. \label{eqn:projection}
\end{equation}
Denoting by $\ve{i}\in\Rm$ the $i$th standard basis vector, $\J^\T \ve{i}$ is the $i$th row of $\J$.
$\upgrad$ is defined as
\begin{equation}
\upgrad(\J)=\frac{1}{m}\sum_{i\in\range{m}} \proj_\J(\J^\T \ve{i}).\label{eqn:upgrad}
\end{equation}

Since the dual cone is convex, it is closed under positive combinations of its elements.
For any $\J$, $\upgrad(\J)$ is thus always in the dual cone of the rows of $\J$, so $\upgrad$ is non-conflicting.
Note that if no pair of gradients conflicts, $\upgrad$ simply averages the rows of the Jacobian.

Since $\pi_\J$ is a projection onto a closed convex cone, if $\vx\in\Rn$ and $0<a\in\mathbb R$, then $\proj_\J(a\cdot \vx)=a \cdot \proj_\J(\vx)$ (see Lemma~\ref{lem:projection_positive_homogeneity}).
Moreover, for any $\0 \plt \vc \in \Rm$, the dual cone of the rows of $\diag(\vc) \cdot \J$ is the same as that of the rows of $\J$.  
By (\ref{eqn:upgrad}), $\upgrad$ is thus linear under scaling.

When $n$ is large, the projection in (\ref{eqn:projection}) is prohibitively expensive to compute.
An alternative but equivalent approach is to use its dual formulation, which is independent of $n$.

\begin{proposition}\label{prop:upgrad_duality}
Let $\J\in\Rmn$.
For any $\vu\in\Rm$, $\proj_\J(\J^\T \vu)=\J^\T \vw$ with
\begin{equation}
\vw\in\argmin_{\substack{\vv\in\Rm:~\vu\pleq \vv}} \vv^\T \J\J^\T \vv.\label{eqn:upgrad_dual}
\end{equation}
\begin{proof}
See Appendix~\ref{appendix:duality_proposition}.
\end{proof}
\end{proposition}

The problem defined in (\ref{eqn:upgrad_dual}) can be solved efficiently using a quadratic programming solver, such as those bundled in \texttt{qpsolvers}~\citep{qpsolvers}.
For any $i\in\range{m}$, let $\vw_i$ be given by (\ref{eqn:upgrad_dual}) when substituting $\vu$ with $\ve{i}$.
Then, by Proposition~\ref{prop:upgrad_duality},
\begin{equation}
\upgrad(\J)=\J^\T \left(\frac{1}{m}\sum_{i\in\range{m}}  \vw_i\right).\label{eqn:upgrad_weights}
\end{equation}
This provides an efficient implementation of $\upgrad$ and proves that it is weighted.
$\upgrad$ can also be easily extended to incorporate a vector of preferences by replacing the average in (\ref{eqn:upgrad}) and (\ref{eqn:upgrad_weights}) by a weighted sum with positive weights.
This extension remains non-conflicting, linear under scaling, and weighted.

\subsection{Convergence to the Pareto front}\label{subsection:convergence}
We now provide theoretical convergence guarantees of JD with $\upgrad$ when minimizing some $\vf:\Rn\to\Rm$ satisfying standard assumptions.
If for a given $\vx\in\Rn$, there exists no $\vy\in\Rn$ satisfying $\vf(\vy)\plneq \vf(\vx)$, then $\vx$ is said to be \emph{Pareto optimal}.
The set $X^\ast\subseteq\Rn$ of Pareto optimal points is called the \emph{Pareto set}, and its image $\vf(X^\ast)$ is called the \emph{Pareto front}.

Whenever $\vf\big(\lambda \vx+(1-\lambda) \vy\big)\pleq \lambda \vf(\vx) + (1-\lambda) \vf(\vy)$ holds for any pair of vectors $\vx,\vy \in\Rn$ and any $\lambda\in [0,1]$, $\vf$ is said to be \emph{$\pleq$-convex}.
Moreover, $\vf$ is said to be \emph{$\beta$-smooth} whenever $\big\| \jac \vf(\vx)-\jac \vf(\vy)\big\|_\Frob\ \leq\beta\|\vx-\vy\|$ holds for any pair of vectors $\vx,\vy\in\Rn$.

\begin{theorem}\label{thm:convergence}
Let $\vf:\Rn\to\Rm$ be a $\beta$-smooth and $\pleq$-convex function.
Suppose that the Pareto front $\vf(X^\ast)$ is compact and that for any $\vx\in\Rn$, there is $\vx^\ast\in X^\ast$ satisfying $\vf(\vx^\ast)\pleq \vf(\vx)$.\footnote{This condition is a generalization to the case $m\geq1$ of the existence of a minimizer $\vx^\ast\in\Rn$ when $m=1$.}
Let $\vx_1\in\Rn$, and for all $t \geq 1$, $\vx_{t+1}=\vx_t - \eta \upgrad\big(\jac \vf(\vx_t)\big)$, with $\eta=\frac{1}{\beta\sqrt{m}}$.
Let $\vw_t$ be the weights defining $\upgrad\big(\jac \vf(\vx_t)\big)$ as per (\ref{eqn:upgrad_weights}), \ie $\upgrad\big(\jac \vf(\vx_t)\big) = \jac \vf(\vx_t)^\T\cdot \vw_t$.
If $\vw_t$ is bounded, then $\vf(\vx_t)$ converges to $\vf(\vx^\ast)$ for some $\vx^\ast\in X^\ast$.
In other words, $\vf(\vx_t)$ converges to the Pareto front.
\begin{proof}[Proof sketch]
The proof follows the standard template used in smooth convex optimization.
The first step is to show that $\vf(\vx_t)$ converges to some limit $\vf^\ast\in\Rm$.
For each $t$, there exists $\vx_t^\ast\in X^\ast$ such that $\vf(\vx_t^\ast)\pleq\vf(\vx_t)$.
By compactness of the Pareto front, a subsequence of $\vf(\vx_t^\ast)$ converges to some $\vf(\vx^\ast)$ with $\vx^\ast\in X^\ast$, satisfying $\vf(\vx^\ast)\pleq\vf^\ast$.
The remainder of the proof shows that this inequality cannot be strict, as this would contradict the convergence of $\vf(\vx_t)$ to $\vf^\ast$.
Hence $\vf(\vx^\ast)=\vf^\ast$.
The full proof is given in Appendix~\ref{appendix:convergence_theorem}.
\end{proof}
\end{theorem}

Another common theoretical result for multi-objective optimization is convergence to a weakly stationary point.
If for a given $\vx\in\Rn$, there exists $\0\plneq \vw$ satisfying $\jac \vf(\vx)^\T \vw = 0$ then $\vx$ is said to be \emph{weakly Pareto stationary}.
Even though every Pareto optimal point is weakly Pareto stationary, the converse does not hold, even in the convex case.
Despite being necessary, convergence to a weakly Pareto stationary point is thus not a sufficient condition for optimality and, hence, constitutes a rather weak guarantee.
To the best of our knowledge, $\upgrad$ is the first non-conflicting aggregator that provably converges to the Pareto front in the smooth convex case.

Appendix~\ref{appendix:supplementary_theory} provides additional results and discussions on the convergence rate and about guarantees in the non-convex setting.

\subsection{Instance-wise risk minimization}\label{subsection:iwrm}
In machine learning, we generally have access to a training set consisting of $m$ examples.
The goal of empirical risk minimization (ERM)~\citep{erm} is simply to minimize the average loss over the whole training set.
More generally, we introduce instance-wise risk minimization (IWRM), which considers the loss associated with each training example as a distinct objective.
Formally, if $\vtheta \in \Rn$ are the parameters of the model and $f_i(\vtheta)$ is the loss associated to the $i$th example, the respective objective functions of ERM and IWRM are:
\begin{align}
&\text{(Empirical risk)} &\bar{f}(\vtheta) &= \frac{1}{m}\sum_{i\in\range{m}} f_i(\vtheta)\\
&\text{(Instance-wise risk)} &\vf(\vtheta) &= \begin{bmatrix} f_1(\vtheta) & f_2(\vtheta) & \cdots & f_m(\vtheta) \end{bmatrix}^\T
\end{align}

Naively using GD for ERM is inefficient in most practical cases, so a prevalent alternative is to use SGD or one of its variants.
Similarly, using JD for IWRM is typically intractable.
Indeed, it would require computing a Jacobian matrix with one row per training example at each iteration.
In contrast, we can use the Jacobian of a random batch of training example losses.
Since it consists of a subset of the rows of the full Jacobian, this approach is a form of stochastic sub-Jacobian descent, as introduced in Section~\ref{subsection:jd}.
IWRM can also be extended to cases where each $f_i$ is a vector-valued function.
The objective would then be the concatenation of the losses of all examples.

We discuss about more standard applications, such as multitask learning, adversarial training and distributed optimization, in Appendix~\ref{appendix:applications}.
\section{Existing aggregators}\label{sec:related}
In the context of multi-task learning, several works have proposed iterative optimization algorithms based on the combination of task-specific gradients~\citep{mgda2,pcgrad,imtl,cagrad,random,nashmtl,aligned_mtl}.
These methods can be formulated as variants of JD parameterized by different aggregators.
More specifically, since the gradients are stochastically estimated from batches of data, these are cases of what we call SEJD.
In the following, we briefly present the most prominent aggregators and summarize their properties in Table~\ref{tab:properties}.
As a baseline, we also consider $\meanagg$, which simply averages the rows of the Jacobian.
Their formal definitions are provided in Appendix~\ref{appendix:properties}.
Some of them are also illustrated in Figure~\ref{fig:other_directions}.

$\rgw$ aggregates the matrix using a random vector of weights~\citep{random}.
$\mgda$ gives the aggregation that maximizes the smallest improvement~\citep{fs, mgda1, mgda2}, which is a non-conflicting update.
$\cagrad$ maximizes the smallest improvement in a ball around the average gradient whose radius is parameterized by $c\in\left[0,1\right[{}$~\citep{cagrad}.
$\pcgrad$ projects each gradient onto the orthogonal hyperplane of other gradients in case of conflict, iteratively and in a random order~\citep{pcgrad}, thereby removing some conflict.
It is, however, only non-conflicting when $m \leq 2$, in which case $\pcgrad = m \cdot \upgrad$.
IMTL-G is a method to balance some gradients with impartiality~\citep{imtl}: it seeks a combination of the gradients having the same projection onto each of their directions.
It is only defined for linearly independent gradients, but we generalize it as a formal aggregator, denoted $\imtlg$, in Appendix~\ref{appendix:imtlg}.
Aligned-MTL orthonormalizes the Jacobian, so that no gradient dominates the others, and weights its rows according to some preferences~\citep{aligned_mtl}.
We denote by $\alignedmtl$ this method with uniform preferences.
$\nashmtl$ aggregates Jacobians by finding the Nash equilibrium between task-specific gradients~\citep{nashmtl}, seen as players bargaining over the update direction.
Lastly, the GradDrop layer~\citep{graddrop} defines a custom backward pass that combines gradients with respect to some internal activation. The corresponding aggregator, denoted $\graddrop$, randomly drops out some gradient coordinates based on their sign and sums the remaining ones.

In the context of continual learning, to limit forgetting, an idea is to project the gradient onto the dual cone of gradients computed with past examples~\citep{gem}.
This idea can be translated into an aggregator that projects the mean gradient onto the dual cone of the rows of the Jacobian, \ie that finds the non-conflicting vector closest to the mean gradient.
We name this $\dualproj$.

Several other works consider the gradients to be noisy when making their theoretical analysis~\citep{crmogm, moco, modo, sdmgrad, stochastic_fs}.
Their solutions for combining gradients are typically stateful.
Although this could enhance practical convergence rates, we have restricted our focus to the analysis of stateless aggregators.
Exploring and analyzing a generalized Jacobian descent algorithm, that would preserve some state over the iterations, is a promising future direction.

In the federated learning setting, several aggregators have been proposed to combine the model updates while being robust to adversaries~\citep{krum, byzantine_gradient_descent, bulyan, trimmed_mean}.
We do not study them here as they mainly focus on security aspects.

\begin{table}[ht!]
\rowcolors{2}{table_odd_row_color}{}
\caption{Properties satisfied for any number of objectives. Proofs are provided in Appendix~\ref{appendix:properties}.}
\label{tab:properties}
\centering
\begin{center}
\begin{tabular}{ll*{3}{c}}
\toprule
Ref. & Aggregator & \makecell[c]{Non-\\conflicting} & \makecell[c]{Linear under\\scaling} & Weighted\\
\midrule
\,\NA & $\meanagg$ & \xmark & \cmark & \cmark\\
\citet{mgda1} & $\mgda$ & \cmark & \xmark & \cmark\\
\citet{gem} & $\dualproj$ & \cmark & \xmark & \cmark\\
\citet{pcgrad} & $\pcgrad$ & \xmark & \cmark & \cmark\\
\citet{graddrop} & $\graddrop$  & \xmark & \xmark & \xmark\\
\citet{imtl} & $\imtlg$ & \xmark & \xmark & \cmark\\
\citet{cagrad} & $\cagrad$ & \xmark  & \xmark & \cmark\\
\citet{random} & $\rgw$ & \xmark & \cmark  & \cmark\\
\citet{nashmtl} & $\nashmtl$ & \cmark & \xmark & \cmark\\
\citet{aligned_mtl} & $\alignedmtl$ & \xmark & \xmark & \cmark\\
(ours) & $\upgrad$ & \cmark & \cmark & \cmark\\
\bottomrule
\end{tabular}
\end{center}
\end{table}
\section{Experiments}\label{sec:experiments}
Appendix~\ref{appendix:optimization_trajectories} shows the optimization trajectories of the methods from Table~\ref{tab:properties} when optimizing three simple convex quadratic functions.
In particular, it confirms the following points: (1) convergence to a weakly Pareto stationary point is insufficient, (2) non-conflicting aggregators do not increase the value of any objective over the course of the optimization, (3) aiming for a particular point of the Pareto front can take much longer than aiming for the Pareto front in general.

In the following, we present empirical results for instance-wise risk minimization on some simple vision datasets.
IWRM is performed by stochastic sub-Jacobian descent, as described in Section~\ref{subsection:iwrm}, with $f_i(\vtheta) = \mathcal{L}_{\mathrm{CE}}\big(\vg(\vx_i; \vtheta), \vy_i\big)$, where $(\vx_i, \vy_i)$ is the $i$'th input/target pair, $\vg(\vx_i; \vtheta)$ is the model output, and $\mathcal{L}_{\mathrm{CE}}$ is the cross-entropy loss.
A key consideration is that when the aggregator is $\meanagg$, this approach becomes equivalent to empirical risk minimization with SGD (indeed, the average gradient is always equal to the gradient of the average objective).
It is thus used as a baseline for comparison.

We train convolutional neural networks for image classification on subsets of SVHN~\citep{svhn}, CIFAR-10~\citep{cifar10}, EuroSAT~\citep{eurosat}, MNIST~\citep{mnist}, Fashion-MNIST~\citep{fashion_mnist} and Kuzushiji-MNIST~\citep{kmnist}.
To make the comparisons as fair as possible, we have tuned the learning rate very precisely for each aggregator, as explained in detail in Appendix \ref{appendix:lr_selection}.
We have also run the same experiments several times independently to gain confidence in our results.
Since this leads to a total of $43776$ training runs across all of our experiments, we have limited the size of each training dataset to 1024 images, greatly reducing computational costs.
Other experimental settings, such as the network architectures and the total computational budget used to run our experiments, are given in Appendix~\ref{appendix:experimental_settings}.
Figure~\ref{fig:cifar10_training_losses} reports the evolution of the training loss on CIFAR-10, one of the datasets with the largest performance gap.
Figure~\ref{fig:cifar10_similarities} shows the similarity between the update of each aggregator and the update given by $\meanagg$ (\ie the update of SGD).

\newcommand{\maintextresultsrow}[2]{
\begin{subfigure}[b]{0.49\textwidth}
\caption{#2: training loss\label{fig:#1_training_losses}}
\includegraphics[width=\textwidth]{images/#1/train_losses_1.pdf}
\end{subfigure}
\begin{subfigure}[b]{0.49\textwidth}
\caption{#2: update similarity to the SGD update\label{fig:#1_similarities}}
\includegraphics[width=\textwidth]{images/#1/similarities_1.pdf}
\end{subfigure}
}

\begin{figure}[ht]
\centering
\maintextresultsrow{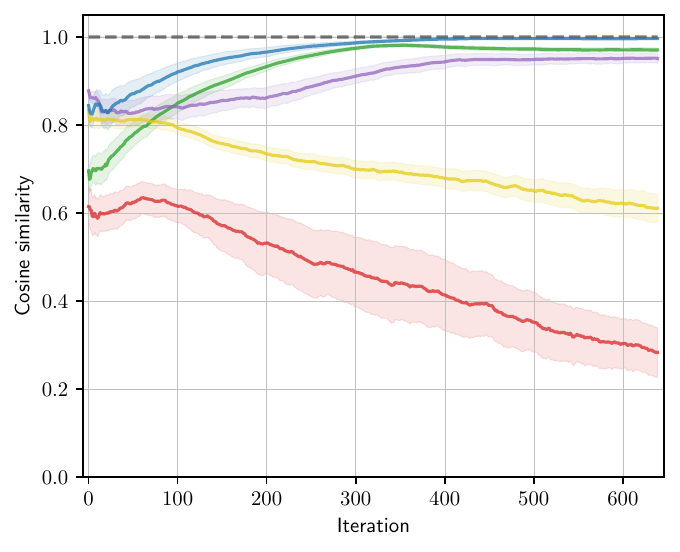}{CIFAR-10}
\caption{Optimization metrics obtained with IWRM with 1024 training examples and a batch size of 32, averaged over \nruns\ independent runs. The shaded area around each curve shows the estimated standard error of the mean over the \nruns\ runs. Curves are smoothed for readability. Best viewed in color.\label{fig:iwrm_results}}
\end{figure}

Here, we compare the aggregators in terms of their average loss over the training set: the goal of ERM.
For this reason, it is rather surprising that $\meanagg$, which directly optimizes this objective, shows a slower convergence rate than $\upgrad$ and, to a lesser extent, $\dualproj$.
This performance gap can be observed on all datasets, as reported in Appendix~\ref{appendix:all_datasets}.

For $\upgrad$, a low similarity with the SGD update indicates that there are some conflicting gradients with imbalanced norms (\ie the setting illustrated in Figure~\ref{fig:aggregator_directions}).
Our interpretation is thus that $\upgrad$ prevents gradients of hard examples from being dominated by those of easier examples early into the training.
Since fitting those is more complex and time-consuming, it could be beneficial to consider them earlier.
Another possible interpretation is that $\upgrad$'s updates could be safer than those of $\meanagg$, allowing it to operate at a higher learning rate.

The sub-optimal performance of $\mgda$ in this setting can be attributed to its sensitivity to small gradients: if any row of the Jacobian approaches zero, the aggregation by $\mgda$ will also approach zero.
This observation illustrates the discrepancy between aiming for weak stationarity and aiming for optimality, as discussed in Section~\ref{subsection:convergence}.
A notable advantage of linearity under scaling is to explicitly prevent this from happening.
Although $\pcgrad$ is also based on gradient projections, its stochastic nature and lack of the non-conflicting property appear to result in instability and significantly worse performance compared to $\upgrad$.

\begin{table}[t]
\rowcolors{2}{table_odd_row_color}{}
\centering
\small
\caption{Test set accuracies (\%) obtained by IWRM with different aggregators across datasets. The corresponding box plots are shown in Figure~\ref{fig:test_results_boxplots}.}
\label{tab:test_results}
\begin{tabular}{lcccccc}
\toprule
Aggregator & SVHN & CIFAR-10 & EuroSAT & MNIST & F-MNIST & K-MNIST \\
\midrule
$\meanagg$        & 69.6 & 47.7 & 68.5 & \textbf{96.3} & 80.7 & 76.9 \\
$\upgrad$  & \textbf{69.7} & \textbf{48.1} & \textbf{73.0} & 96.0 & 80.8 & \textbf{77.1} \\
$\mgda$           & 41.7 & 32.9 & 62.5 & 85.2 & 73.4 & 53.7 \\
$\alignedmtl$     & 61.3 & 47.3 & 65.3 & 88.3 & 78.5 & 67.9 \\
$\pcgrad$         & 49.1 & \textbf{48.1} & 59.3 & 85.9 & 69.3 & 54.7 \\
$\dualproj$       & 69.3 & 47.2 & 72.7 & 96.1 & 80.8 & 76.4 \\
$\rgw$            & 66.2 & 39.5 & 67.9 & 94.0 & 77.1 & 73.6 \\
$\graddrop$       & 68.9 & 46.8 & 71.9 & \textbf{96.3} & \textbf{80.9} & 76.9 \\
$\cagrad$         & 60.7 & 42.4 & 68.1 & 90.3 & 78.1 & 65.3 \\
$\imtlg$          & 58.7 & 44.8 & 65.2 & 87.2 & 75.0 & 65.9 \\
$\nashmtl$        & 39.7 & 31.3 & 60.8 & 87.6 & 72.0 & 52.7 \\
\bottomrule
\end{tabular}
\end{table}

Table~\ref{tab:test_results} reports the test set accuracies, showing that IWRM with $\upgrad$ can also lead to a better generalization.
While increasing the batch size in SGD reduces variance, the effect of doing so in SSJD combined with $\upgrad$ is non-trivial, as it also tightens the dual cone.
Additional results obtained when varying the batch size or updating the parameters with the \texttt{Adam} optimizer are available in Appendices~\ref{appendix:batch_size_study} and~\ref{appendix:adam_study}, respectively.
Appendix~\ref{appendix:speed} provides memory usage and computation time considerations for our methods.

This experiment provides an example of a setting, that is traditionally considered as single-objective, benefitting from a multi-objective formulation.
Even though more work is required to reduce IWRM's overhead and to validate it on more realistic learning tasks, we believe that it deserves further attention.
\section{Gramian-based Jacobian descent}\label{sec:implementation}
When the number of objectives is dominated by the number of parameters of the model, the main overhead of JD comes from storing the Jacobian matrix, $m$ times larger than a single gradient, rather than the gradient itself.
In the following, we introduce an alternative implementation of JD that only requires the $m (m+1) / 2$ inner products between each pair of gradients.

For any $\J \in \Rmn$, the matrix $\G = \J\J^\T$ is called the \emph{Gramian} of $\J$ and is positive semi-definite.
Let $\PSD{m}\subseteq \mathbb R^{m\times m}$ be the set of positive semi-definite matrices.
The Gramian of the Jacobian, denoted $\gram \vf(\vx) = \jac \vf(\vx) \cdot \jac \vf(\vx)^\T \in \PSD{m}$, contains all the inner products between pairs of gradients.
It thus includes the gradient norms, and captures the conflict between them.
Whenever $\A$ is a weighted aggregator, the update of JD is $\vy=-\eta\jac \vf(\vx)^\T \vw$ for some vector of weights $\vw\in\Rm$.
Substituting this into the Taylor approximation of (\ref{eqn:taylor-approximation}) gives (see Appendix~\ref{appendix:gramian_based_derivations})
\begin{equation}
\vf(\vx+\vy)=\vf(\vx)-\eta \gram \vf(\vx)\cdot \vw + o\left(\eta\sqrt{\vw^\T\cdot \gram \vf(\vx) \cdot \vw}\right).\label{eqn:taylor-approxmiation-gramian}
\end{equation}
This expression only depends on the Jacobian through its Gramian.
It is thus sensible to focus on aggregators whose weights are only a function of the Gramian.
Denoting this function as $\W~:~\PSD{m}\to\Rm$, those aggregators satisfy $\A(\J) = \J^\T\cdot \W(\G)$.
Remarkably, all weighted aggregators of Table~\ref{tab:properties} can be expressed in this form. In the case of $\upgrad$, this is clearly demonstrated in Proposition~\ref{prop:upgrad_duality}, which shows that the weights depend on $\G$.
For such aggregators, substitution and linearity of differentiation then yield (see Appendix~\ref{appendix:gramian_based_derivations}), with $\W\big(\gram \vf(\vx)\big)$ treated as a constant during differentiation,
\begin{equation}
\A\big(\jac \vf(\vx)\big) = \nabla \Big(\W\big(\gram \vf(\vx)\big)^\T \cdot \vf\Big)(\vx).\label{eqn:gramian_based_update}
\end{equation}
After computing $\W\big(\gram \vf(\vx)\big)$, a step of JD would thus only require the backpropagation of a scalar function.
The computational cost of applying $\W$ depends on the aggregator and is often dominated by the cost of computing the Gramian.

We now outline a method to compute the Gramian of the Jacobian without ever having to store the full Jacobian in memory.
Similarly to the backpropagation algorithm, we can use the chain rule.
Let $\vg:\Rn \to \Rk$ and $\vf:\Rk\to\Rm$, then for any $\vx\in\Rn$, the chain rule for Gramians is
\begin{align}
\gram (\vf\circ \vg)(\vx)=\jac \vf\big(\vg(\vx)\big) \cdot \gram \vg(\vx) \cdot  \jac \vf\big(\vg(\vx)\big)^\T.\label{eqn:gramian_chain_rule}
\end{align}
Moreover, when the function has multiple inputs, the Gramian can be computed as a sum of individual Gramians.
Let $\vf:\mathbb R^{n_1+\cdots+n_k}\to\Rm$ and $\vx = \begin{bmatrix}\vx_1^\T & \cdots & \vx_k^\T\end{bmatrix}^\T$.
We can write $\jac \vf(\vx)$ as the concatenation of Jacobians $\begin{bmatrix} \jac_{\vx_1} \vf(\vx)&\cdots & \jac_{\vx_k} \vf(\vx) \end{bmatrix}$, where $\jac_{\vx_i} \vf(\vx)$ is the Jacobian of $\vf$ with respect to $\vx_i$ evaluated at $\vx$.
For any $i\in\range{k}$, let $\gram_{\vx_i} \vf(\vx)=\jac_{\vx_i} \vf(\vx) \cdot \jac_{\vx_i} \vf(\vx)^\T$.
Then
\begin{align}
\gram \vf(\vx_1, \dots, \vx_k)=\sum_{i\in\range{k}} \gram_{\vx_i} \vf(\vx_1,\dots, \vx_k).\label{eqn:gramian_sum}
\end{align}
When a function is made of compositions and concatenations of elementary functions, the Gramian of the Jacobian can thus be expressed with sums and products of partial Jacobians.

We now provide an example algorithm to compute the Gramian of a sequence of layers.
For $0\leq i < k$, let $\vf_i:\R^{n_i}\times\R^{\ell_i}\to\R^{n_{i+1}}$ be a layer parameterized by $\vp_i \in \R^{\ell_i}$.
Given $\vx_0\in \R^{n_0}$, for $0\leq i < k$, the activations are recursively defined as $\vx_{i+1} = \vf_i(\vx_i, \vp_i)$.
Algorithm~\ref{alg:reverse_accumulation_gramian} illustrates how (\ref{eqn:gramian_chain_rule}) and (\ref{eqn:gramian_sum}) can be combined to compute the Gramian of the network with respect to its parameters.

\begin{algorithm}[H]
\label{alg:reverse_accumulation_gramian}
\caption{\emph{Gramian reverse accumulation} for a sequence of layers}
\DontPrintSemicolon
$\J_x \gets I$ \Comment{Identity matrix of size $n_k\times n_k$}
$\G \hspace{0.09cm}\gets 0$ \hspace{0.02cm}\Comment{Zero matrix of size $n_k\times n_k$}
\For{$i \gets k-1$ \KwTo $0$}{
    $\J_{p} \gets J_x \cdot \jac_{\vp_i}\vf_i(\vx_i, \vp_i)$\hspace{0.03cm}\Comment{Jacobian of $\vx_k$ \wrt $\vp_i$}
    $\G \hspace{0.09cm}\gets \G + \J_{p}\J_{p}^\T$\\
    $\J_x \gets J_x \cdot \jac_{\vx_i}\vf_i(\vx_i, \vp_i)$\Comment{Jacobian of $\vx_k$ \wrt $\vx_i$}
}
\KwOut{$\G$}
\end{algorithm}

Algorithm~\ref{alg:reverse_accumulation_gramian} is analogous to the backpropagation of a gradient, except that it backpropagates the Jacobian $\J_x$ of $\vx_k$ with respect to the current activation, so vector-Jacobian products become Jacobian-Jacobian products.
Whenever the Jacobian $\J_p$ with respect to the parameters of a layer is obtained, its Gramian is accumulated into $\G$, as justified by (\ref{eqn:gramian_sum}), and $\J_p$ can then be freed, so the full Jacobian is never stored.

Our library provides a working implementation of Algorithm~\ref{alg:reverse_accumulation_gramian} supporting any computational graph (rather than just sequential models).
Note that it also provides a Gramian-based weighting implementation for most weighted aggregators, to make them compatible with this algorithm.

The practical memory usage and computation time obtained with this algorithm on the IWRM experiment described in Section~\ref{sec:experiments} are provided in Appendix~\ref{appendix:speed}.
We can see that SSJD with $\upgrad$ becomes between $42\%$ and $89\%$ faster with the Gramian-based implementation compared to the naive implementation, despite being mathematically equivalent.
This implementation also leads to a reduction of the memory usage between $87\%$ and $93\%$ depending on the architecture.
It remains an open challenge to make this method even more efficient, and in particular to make it handle the block-diagonal sparsity of $\J_x$ when using it for IWRM.

\section{Conclusion}
We introduced Jacobian descent and its stochastic variants, together with a novel non-conflicting aggregator that provides stronger convergence guarantees than prior approaches.
Through a Gramian-based implementation, we significantly reduced the memory and time overhead of existing methods.
Combined with our library, these contributions should provide a stronger conceptual foundation for future works and largely improve the practical usability of such methods.

\paragraph{Limitations and future directions.}
The empirical evaluation of Jacobian descent is limited to synthetic functions and to IWRM, a setting with moderately conflicting objectives.
An essential future direction is thus to develop proper benchmarks to compare aggregators on a wider variety of real learning tasks.
Also, the efficiency of SSJD on the IWRM objective, and of $\upgrad$ in general, can still be largely improved through the development of efficient block-diagonal sparse tensors and of a GPU-based quadratic program solver, respectively.
\section*{Acknowledgments}
We would like to express our sincere thanks to Scott Pesme, Emre Telatar, Matthieu Buot de l'\'Epine, Adrien Vandenbroucque, Ye Zhu, Alix Jeannerot, and Damian Dudzicz for their careful and thorough review.
The many insightful discussions that we shared with them were essential to this project.
We are also grateful to Khush Patel, Praneth Paruchuri, and Rasa Khosrowshahi, our most dedicated open-source contributors to TorchJD.

This work received no external funding. The authors declare no competing interests.

\bibliography{content/Extra-bibliography}

\newpage
\appendix

\renewcommand\qedsymbol{$\square$}

\newcommand{\trajectoryheight}{0.27\textheight}
\newcommand{\resultsrow}[2]{
\begin{subfigure}[b]{0.49\textwidth}
\centering
\caption{Training loss}
\includegraphics[width=\textwidth]{images/#1/train_losses_#2.pdf}
\end{subfigure}
\begin{subfigure}[b]{0.49\textwidth}
\centering
\caption{Update similarity to the SGD update}
\includegraphics[width=\textwidth]{images/#1/similarities_#2.pdf}
\end{subfigure}
}
\newcommand{\results}[2]{
\begin{figure}
\centering
\resultsrow{#1}{1}
\resultsrow{#1}{2}
\resultsrow{#1}{3}
\caption{#2 results.\label{fig:#1_additional_results}}
\end{figure}
}
\newcommand{\bssresultsrow}[1]{
\begin{subfigure}[b]{0.49\textwidth}
\centering
\caption{$\text{BS}=#1$: Training loss}
\includegraphics[width=\textwidth]{images/batch_size_study/train_losses_#1.pdf}
\end{subfigure}
\begin{subfigure}[b]{0.49\textwidth}
\centering
\caption{$\text{BS}=#1$: Update similarity to the SGD update}
\includegraphics[width=\textwidth]{images/batch_size_study/similarities_#1.pdf}
\end{subfigure}
}
\newcommand{\bssresults}{
\begin{figure}
\centering
\bssresultsrow{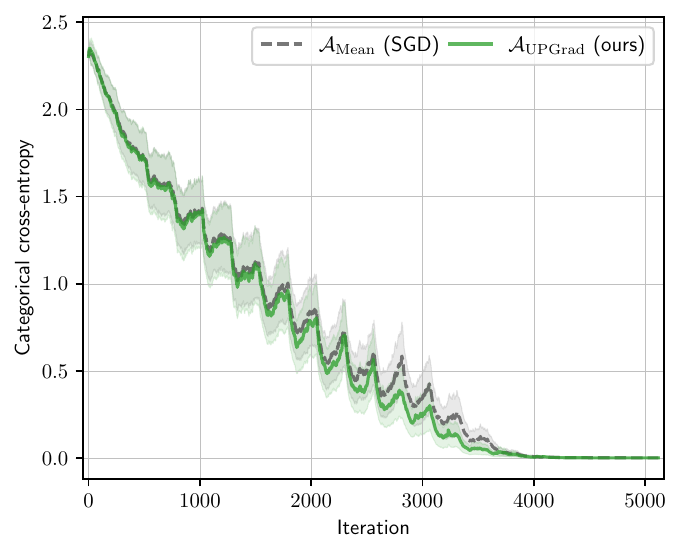}
\bssresultsrow{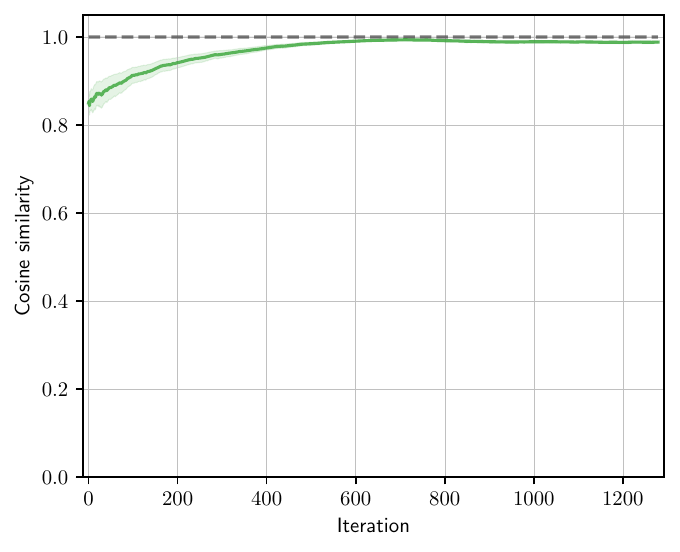}
\bssresultsrow{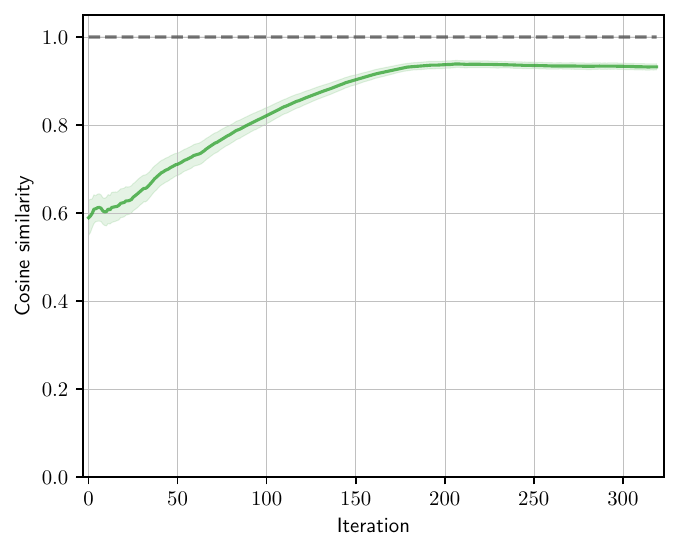}
\caption{CIFAR-10 results with different batch sizes (BS). The number of epochs is always 20, so the number of iterations varies.\label{fig:batch_size_study_results}}
\end{figure}
}
\newcommand{\adamresultsrow}[2]{
\begin{subfigure}[b]{0.49\textwidth}
\centering
\caption{#2: Training loss}
\includegraphics[width=\textwidth]{images/adam_study/train_losses_#1.pdf}
\end{subfigure}
\begin{subfigure}[b]{0.49\textwidth}
\centering
\caption{#2: Update similarity to the SGD update}
\includegraphics[width=\textwidth]{images/adam_study/similarities_#1.pdf}
\end{subfigure}
}
\newcommand{\adamresults}{
\begin{figure}
\centering
\adamresultsrow{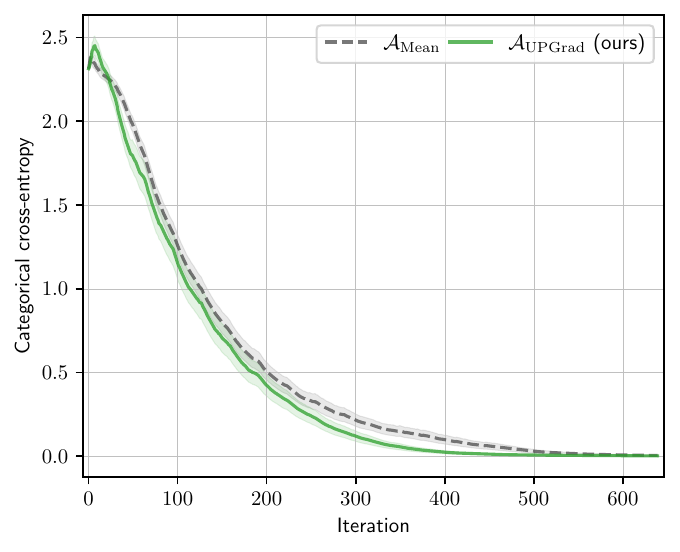}{SVHN}
\adamresultsrow{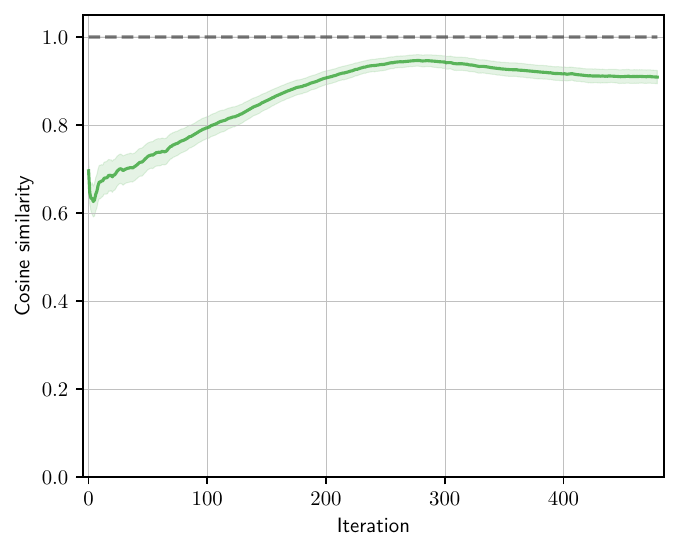}{CIFAR-10}
\caption{Results with the \texttt{Adam} optimizer.\label{fig:adam_study_results}}
\end{figure}
}

\section{Additional experimental results}\label{appendix:additional_results}
In this appendix, we provide additional experimental results about Jacobian descent and IWRM.

\subsection{Optimization trajectories}\label{appendix:optimization_trajectories}
\begin{figure}
\centering
\begin{subfigure}[b]{\textwidth}
\centering
\caption{Parameter space. Black star: Pareto set. Dashed lines: contour lines of the mean objective.\label{fig:ewq_param_trajectories}}
\includegraphics[width=\textwidth,height=\trajectoryheight,keepaspectratio]{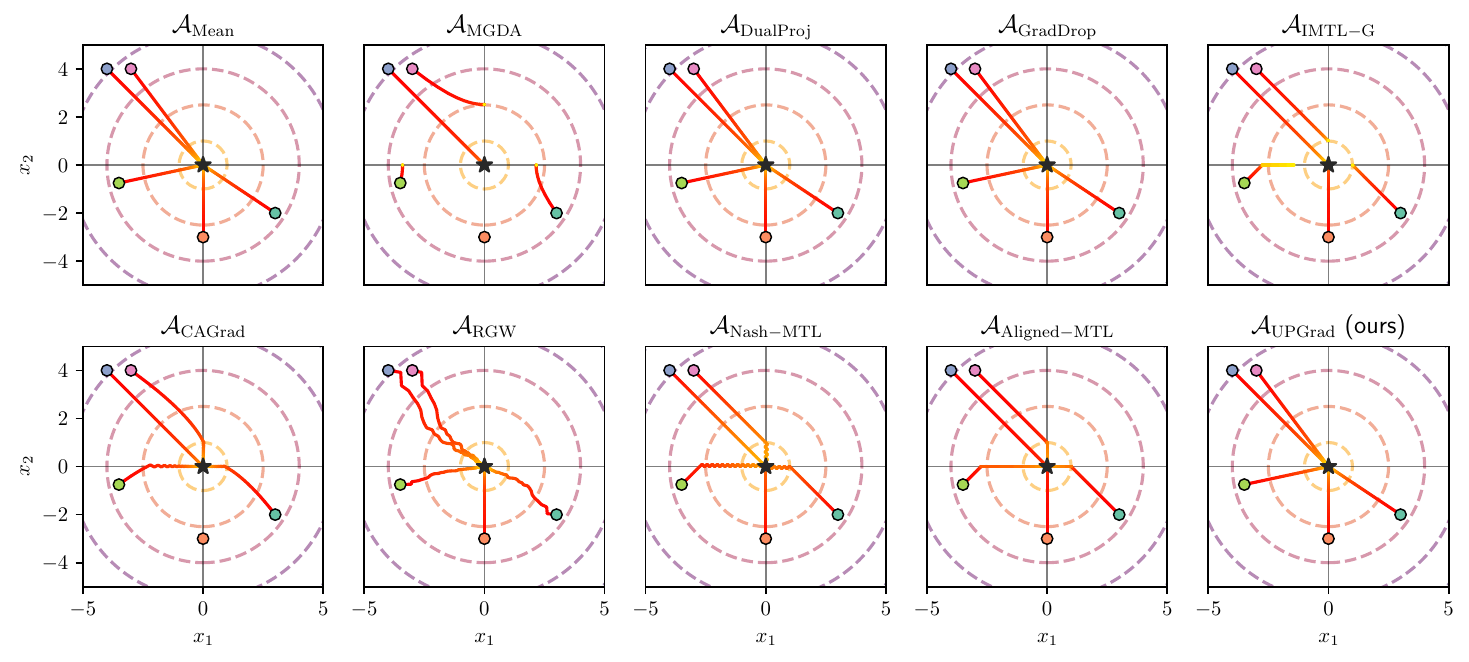}
\end{subfigure}
\begin{subfigure}[b]{\textwidth}
\centering
\caption{Objective space. Black star: Pareto front. Background color: distance to the Pareto front.\label{fig:ewq_value_trajectories}}
\includegraphics[width=\textwidth,height=\trajectoryheight,keepaspectratio]{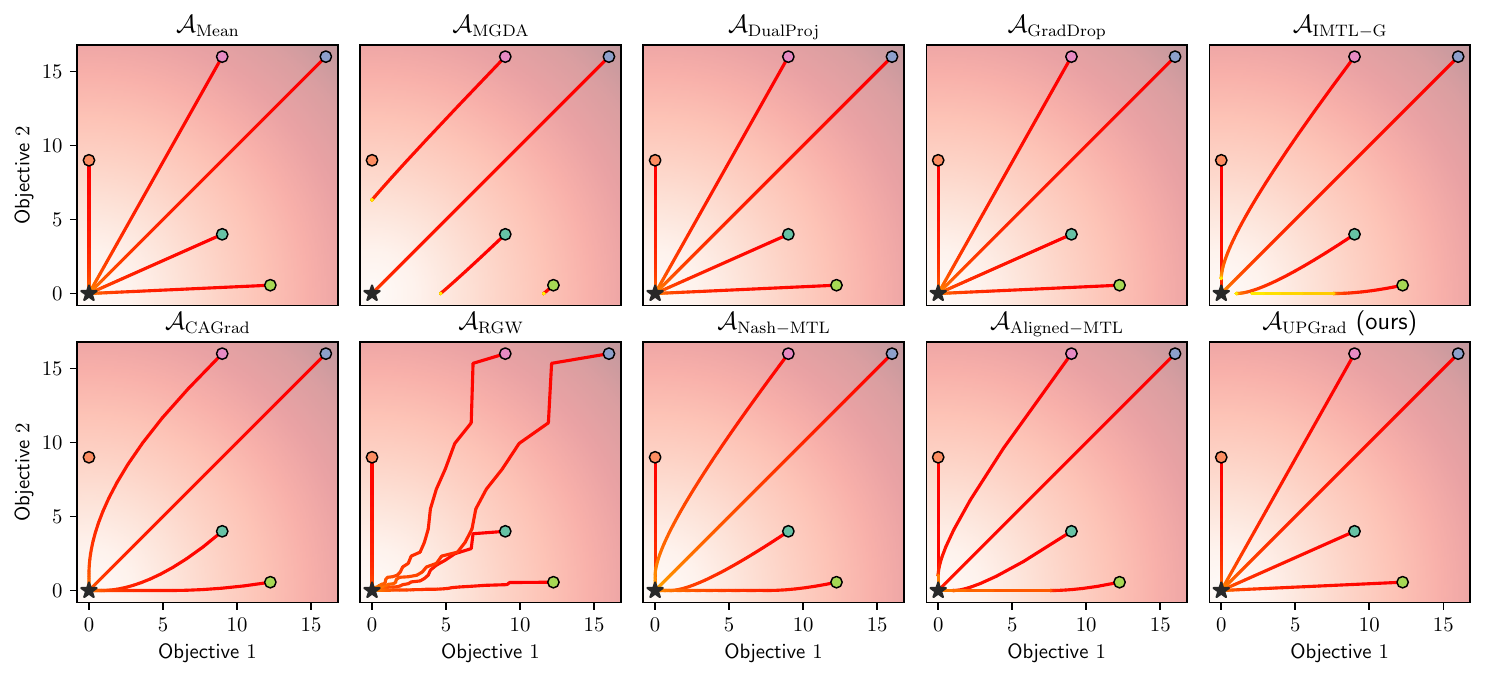}
\end{subfigure}
\begin{subfigure}[b]{\textwidth}
\centering
\caption{Distance to Pareto front over the iterations for each initial point. Each line has the color of the corresponding initial point in the trajectory plots.\label{fig:ewq_distance_to_pf}}
\includegraphics[width=\textwidth,height=\trajectoryheight,keepaspectratio]{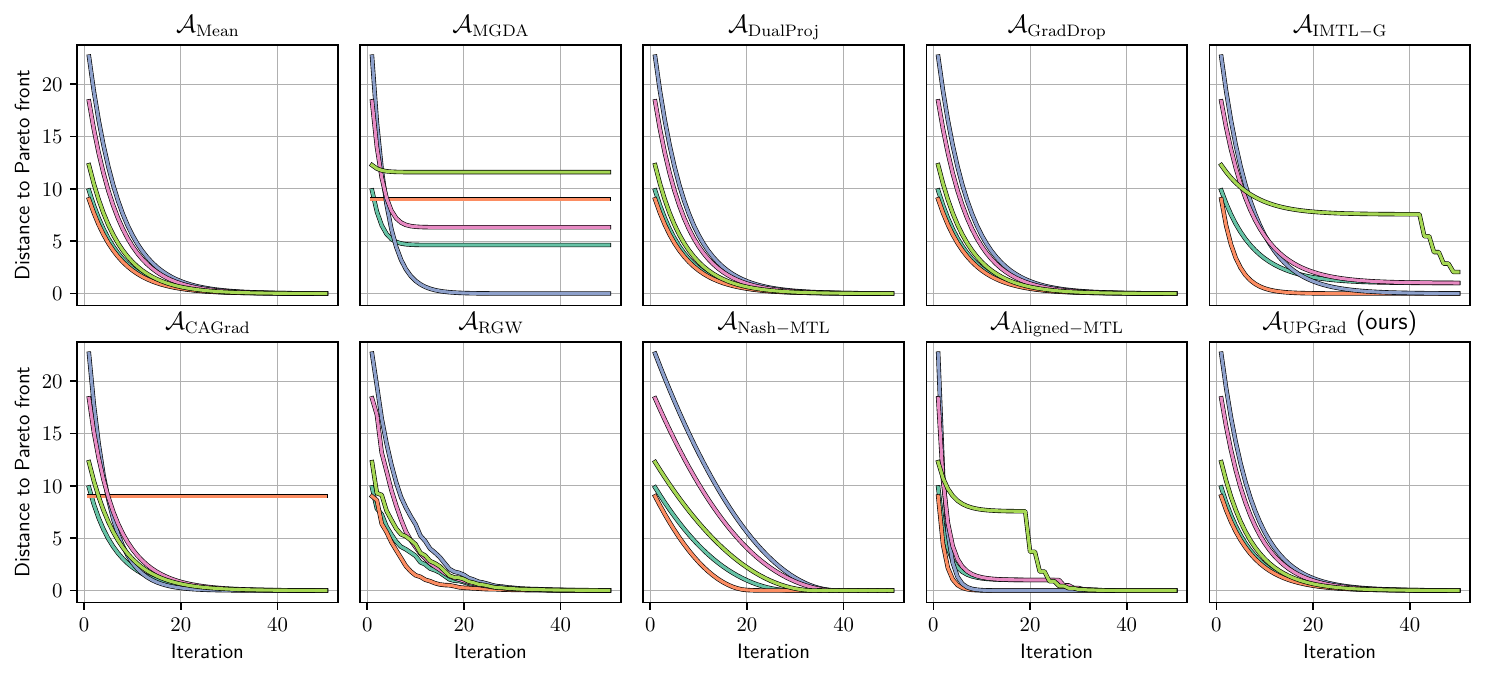}
\end{subfigure}
\caption{Optimization trajectories of various aggregators when optimizing $\vf:\begin{bmatrix} x_1 & x_2 \end{bmatrix}^\T\mapsto \begin{bmatrix} x_1^2 & x_2^2 \end{bmatrix}^\T$ with JD. Colored dots represent the initial points. The trajectories start in red and evolve towards yellow.\label{fig:ewq_trajectories}}
\end{figure}

\begin{figure}
\centering
\begin{subfigure}[b]{\textwidth}
\centering
\caption{Parameter space. Black line: Pareto set. Background color: cosine similarity between the two gradients, between $-1$ (fully conflicting) in pink and $1$ (fully aligned) in green. For instance, the similarity is $-0.925$ at the initial point $\begin{bmatrix} 0 & 0 \end{bmatrix}^\T$ and is $-1$ on (and only on) the Pareto set.\label{fig:cqf_param_trajectories}}
\includegraphics[width=\textwidth,height=\trajectoryheight,keepaspectratio]{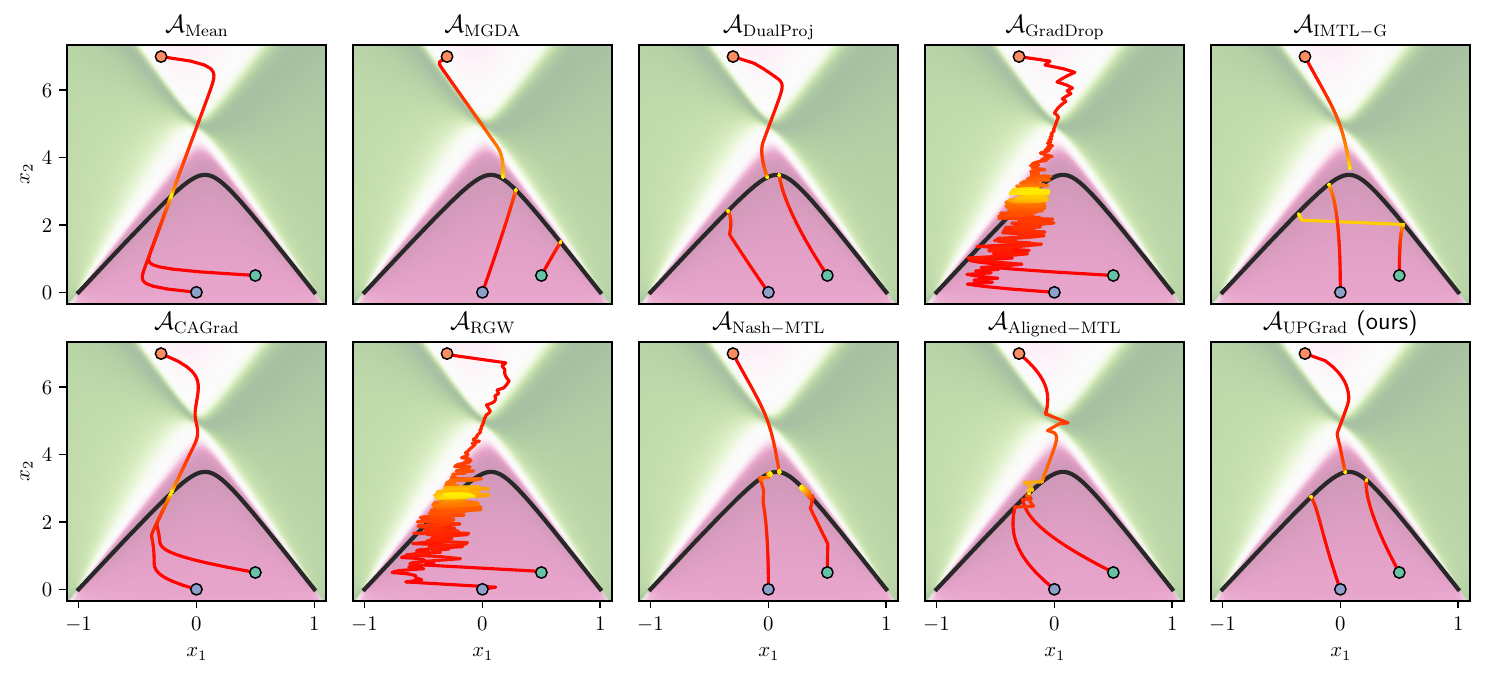}
\end{subfigure}
\begin{subfigure}[b]{\textwidth}
\centering
\caption{Objective space. Black line: Pareto front. Background color: distance to the Pareto front.\label{fig:cqf_value_trajectories}}
\includegraphics[width=\textwidth,height=\trajectoryheight,keepaspectratio]{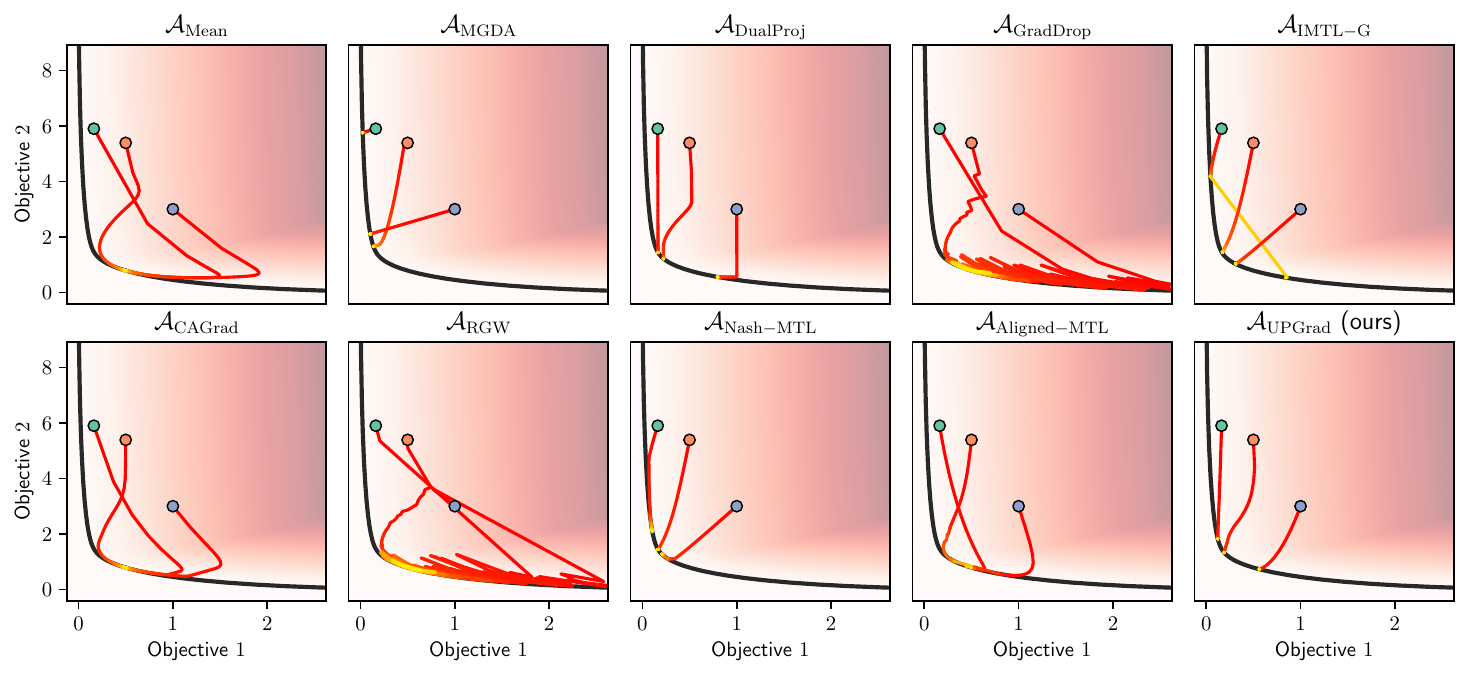}
\end{subfigure}
\begin{subfigure}[b]{\textwidth}
\centering
\caption{Distance to Pareto front over the iterations for each initial point.\label{fig:cqf_distance_to_pf}}
\includegraphics[width=\textwidth,height=\trajectoryheight,keepaspectratio]{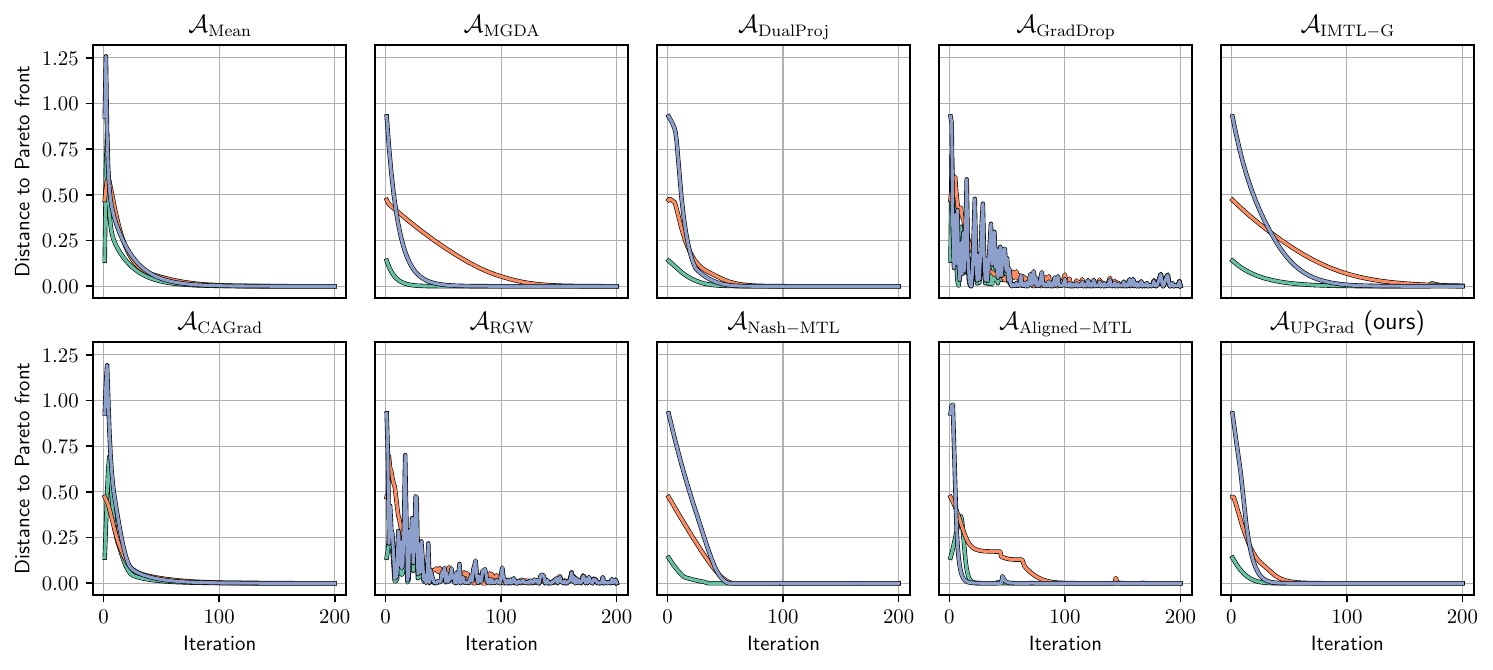}
\end{subfigure}
\caption{Optimization trajectories of various aggregators when optimizing the convex quadratic function $\vg$ of (\ref{eqn:f_cqf}) with JD. Colored dots represent initial parameter values.\label{fig:cqf_trajectories}}
\end{figure}

\begin{figure}
\centering
\begin{subfigure}[b]{\textwidth}
\centering
\caption{Parameter space. Black line: Pareto set. Background color: cosine similarity between the two gradients, between $-1$ (fully conflicting) in pink and $1$ (fully aligned) in green.\label{fig:hqf_param_trajectories}}
\includegraphics[width=\textwidth,height=\trajectoryheight,keepaspectratio]{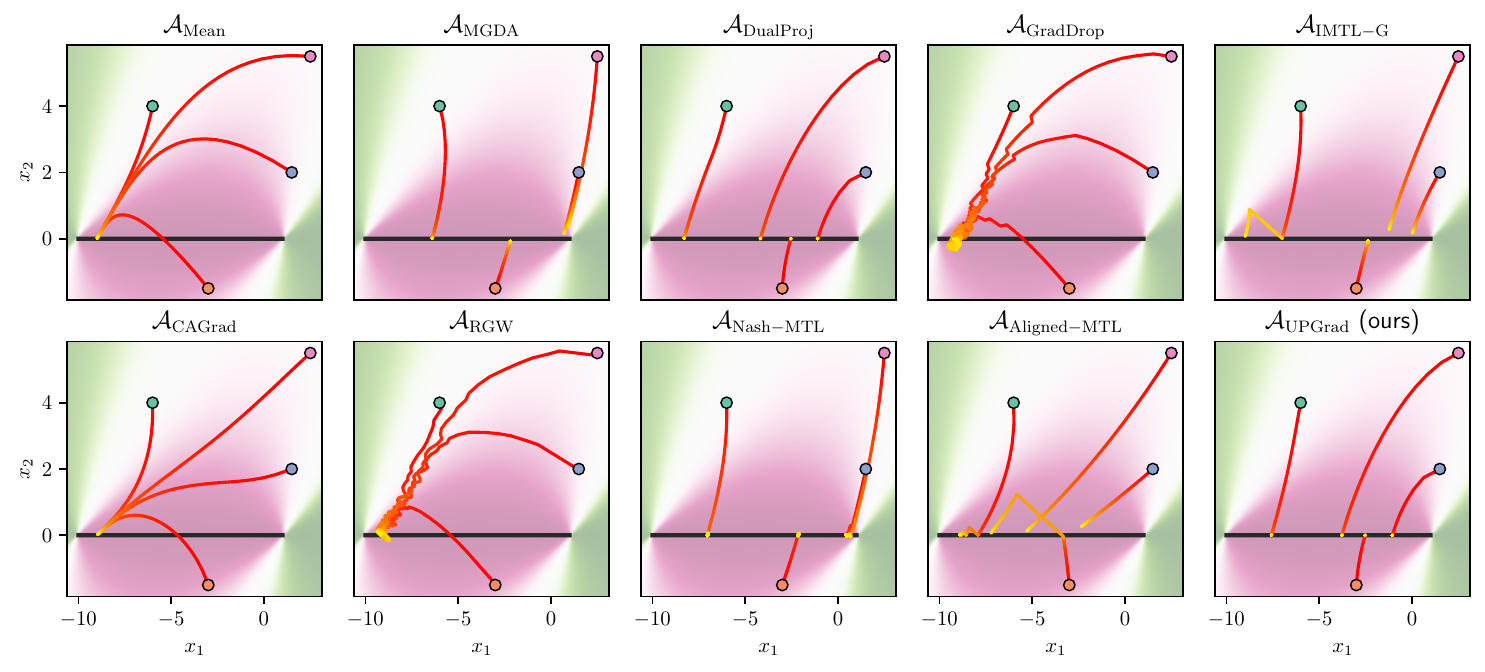}
\end{subfigure}
\begin{subfigure}[b]{\textwidth}
\centering
\caption{Objective space. Black line: Pareto front. Background color: distance to the Pareto front.\label{fig:hqf_value_trajectories}}
\includegraphics[width=\textwidth,height=\trajectoryheight,keepaspectratio]{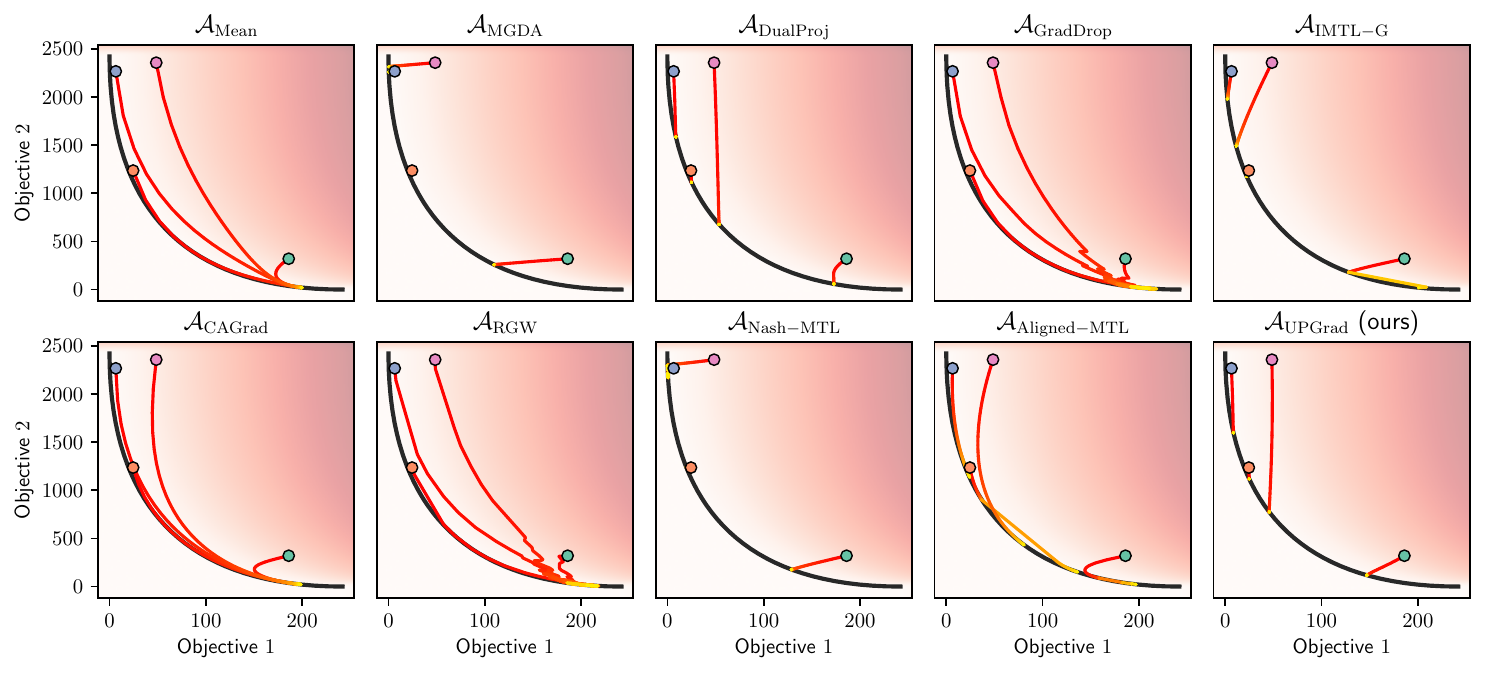}
\end{subfigure}
\begin{subfigure}[b]{\textwidth}
\centering
\caption{Distance to Pareto front over the iterations for each initial point. Each line has the color of the corresponding initial point in the trajectory plots.\label{fig:hqf_distance_to_pf}}
\includegraphics[width=\textwidth,height=\trajectoryheight,keepaspectratio]{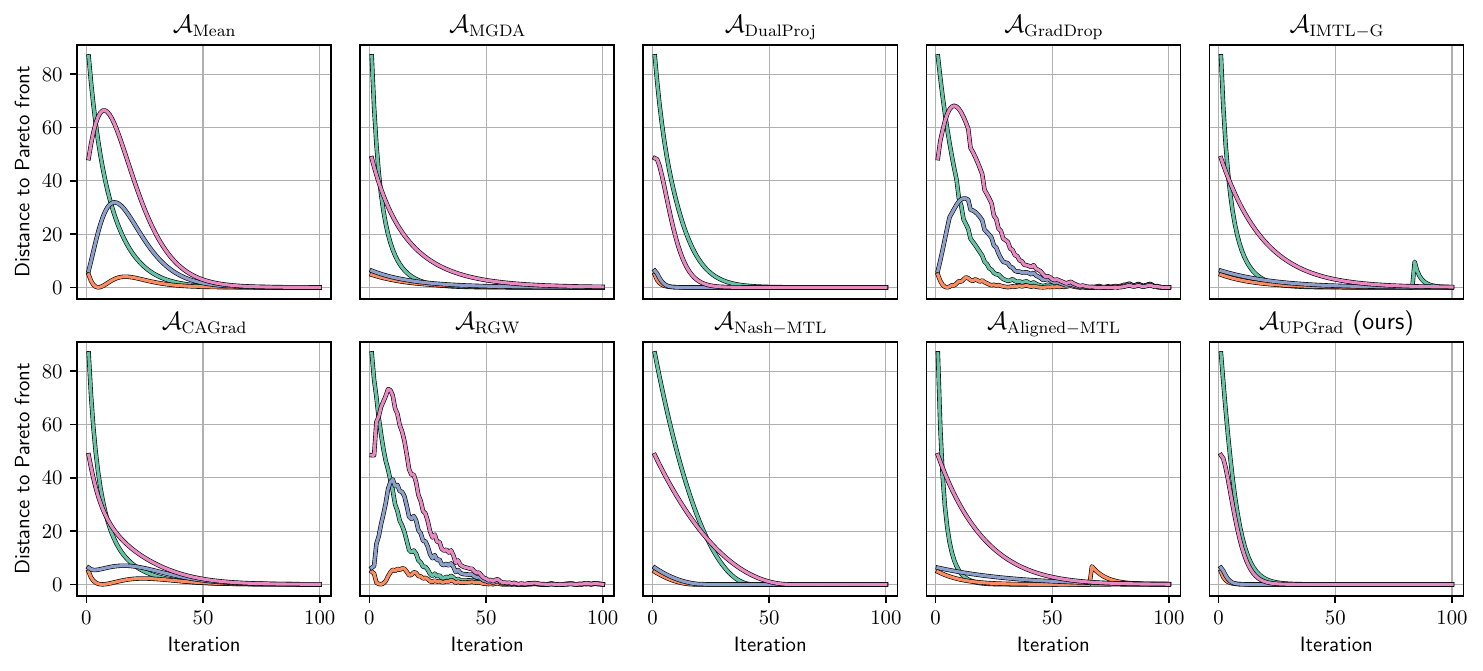}
\end{subfigure}
\caption{Optimization trajectories of various aggregators when optimizing the convex quadratic function $\vh$ of (\ref{eqn:f_hqf}) with JD. Colored dots represent initial parameter values. The trajectories start in red and evolve towards yellow.\label{fig:hqf_trajectories}}
\end{figure}

Figures~\ref{fig:ewq_trajectories}, \ref{fig:cqf_trajectories} and~\ref{fig:hqf_trajectories} show the optimization trajectories of the aggregators of Table~\ref{tab:properties} for three different convex quadratic functions, with several initializations.\footnote{We do not show the trajectories of $\pcgrad$ because it is equivalent to $2 \cdot \upgrad$ when there are only two objectives.}

The first function is the element-wise quadratic function $\vf_\mathrm{EWQ}: \R^2 \to \R^2$, defined as
\begin{equation}
\vf(\vx) = \begin{bmatrix} x_1^2 & x_2^2 \end{bmatrix}^\T.
\end{equation}
Notably, its Pareto set only contains the origin, while its set of weakly Pareto stationary points is the union of the two axes $x_1=0$ and $x_2=0$.
The Jacobian of $\vf$ at $\begin{bmatrix} x_1 \\ x_2 \end{bmatrix}$ is $\begin{bmatrix} 2x_1 & 0\\ 0 & 2x_2 \end{bmatrix}$, indicating that the gradients never conflict, which makes this function relatively simple to optimize.
Nevertheless, Figure~\ref{fig:ewq_trajectories} shows that $\mgda$, $\imtlg$ and $\cagrad$ fail to converge to the Pareto set for some initializations.
For instance, $\mgda$ converges to weakly Pareto stationary points, which can be arbitrarily far from the Pareto set.
This trivial example shows the importance of proving convergence to the Pareto front (as we do for $\upgrad$ in Section~\ref{subsection:convergence}) rather than to a weakly Pareto stationary point.

The second function is a convex quadratic function $\vg: \R^2 \to \R^2$, constructed to sometimes reproduce the setting of Figure~\ref{fig:aggregator_directions} with conflicting and imbalanced gradients:

\begin{align}
\vg\left(\vx\right) &= \begin{bmatrix}(\vx - \vv_1)^\T \cdot A_1 \cdot (\vx - \vv_1) \\ (\vx - \vv_2)^\T \cdot A_2 \cdot (\vx - \vv_2) \end{bmatrix}\label{eqn:f_cqf}\\
\text{with }A_1 &= \begin{bmatrix} 1 & 0.2 \\ 0.2 & 0.05 \end{bmatrix}\text{, }A_2 = \begin{bmatrix} 3 & -0.6 \\ -0.6 & 0.2 \end{bmatrix}\\
\vv_1 &= \begin{bmatrix} 1 \\ 0 \end{bmatrix}\text{, and }\vv_2 = \begin{bmatrix} -1 \\ 0 \end{bmatrix}
\end{align}

Figure~\ref{fig:cqf_trajectories} shows that most existing aggregators have some unusual behavior in this setting. In particular, on Figure~\ref{fig:cqf_value_trajectories}, we can see that with $\meanagg$, $\graddrop$, $\imtlg$, $\cagrad$, $\rgw$ and $\alignedmtl$, the updates can make one objective's value increase.
In contrast, with the non-conflicting aggregators $\mgda$, $\dualproj$, $\nashmtl$ and $\upgrad$, no objective value increases during the optimization.
Due to its linearity under scaling, $\upgrad$ also shows improved smoothness compared to the other non-conflicting aggregators.

The third function, $\vh: \R^2 \to \R^2$, is an even simpler convex quadratic function with conflicting gradients, but with even greater gradient norm imbalance.

\begin{align}
\vh\left(\vx\right) &= \begin{bmatrix}(\vx - \vu_1)^\T \cdot A \cdot (\vx - \vu_1) \\ (\vx - \vu_2)^\T \cdot 10A \cdot (\vx - \vu_2) \end{bmatrix}\label{eqn:f_hqf}\\
\text{with }A &= \begin{bmatrix} 2 & -1 \\ -1 & 2 \end{bmatrix}\\
\vu_1 &= \begin{bmatrix} 1 \\ 0 \end{bmatrix}\text{, and }\vu_2 = \begin{bmatrix} -10 \\ 0 \end{bmatrix}
\end{align}

Figure~\ref{fig:hqf_trajectories} shows that in such a setting, $\meanagg$ can take a much longer trajectory to reach the Pareto front compared to $\upgrad$.
This shows that the advantage of aiming for the Pareto front in general rather than one of its points, can sometimes be very large.
Note that the gap between $\meanagg$ and $\upgrad$ can be arbitrarily large if the gradient imbalance increases.

\newpage

\subsection{All datasets and all aggregators}\label{appendix:all_datasets}

Figures~\ref{fig:svhn_additional_results}, \ref{fig:cifar10_additional_results}, \ref{fig:eurosat_additional_results}, \ref{fig:mnist_additional_results}, \ref{fig:fmnist_additional_results} and~\ref{fig:kmnist_additional_results} show the full results of the experiments described in Section~\ref{sec:experiments} on SVHN, CIFAR-10, EuroSAT, MNIST, Fashion-MNIST and Kuzushiji-MNIST, respectively.
For readability, the results are displayed on three different plots for each dataset.
We always show $\upgrad$ and $\meanagg$ for reference.
The exact experimental settings are described in Appendix~\ref{appendix:experimental_settings}.

It should be noted that some of these aggregators were not developed as general-purpose aggregators, but mainly for the use case of multi-task learning, with one gradient per task.
Our experiments present a more challenging setting than multi-task learning optimization because conflict between rows of the Jacobian is typically higher.

We can see that $\upgrad$ provides a significant improvement over $\meanagg$ on all datasets.
Moreover, the performance gaps seem to be linked to the difficulty of the dataset, which suggests that experimenting with harder tasks is a promising future direction.

\results{svhn}{SVHN}
\results{cifar10}{CIFAR-10}
\results{eurosat}{EuroSAT}
\results{mnist}{MNIST}
\results{fmnist}{Fashion-MNIST}
\results{kmnist}{Kuzushiji-MNIST}

\subsection{Test set accuracies}\label{appendix:test_results}

Figure~\ref{fig:test_results_boxplots} shows the distribution, over the \nruns\ independent runs described in Appendix~\ref{appendix:random_reruns}, of the test set accuracies reported in Table~\ref{tab:test_results}, with one subfigure per dataset.
This makes it easier to compare aggregators and to assess the significance of the differences between them.

\begin{figure}[h]
\centering
\includegraphics[width=\textwidth]{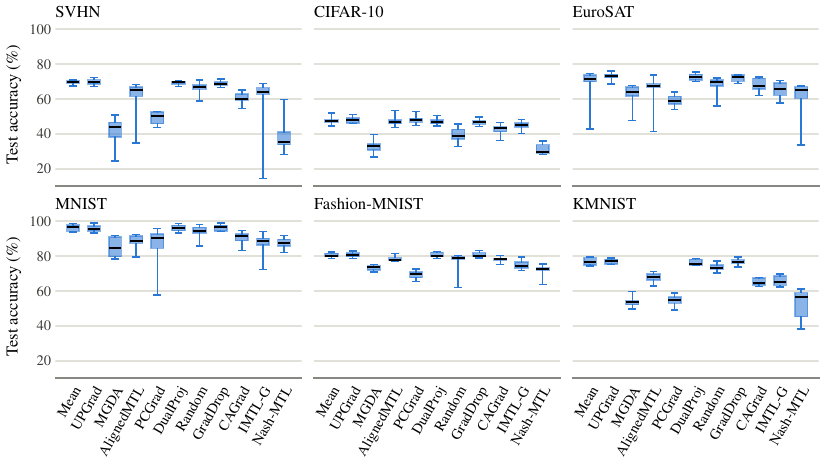}
\caption{Box plots of the test set accuracies (\%) obtained by IWRM with different aggregators across datasets, over \nruns\ independent runs.\label{fig:test_results_boxplots}}
\end{figure}

\subsection{Varying the batch size}\label{appendix:batch_size_study}

Figure~\ref{fig:batch_size_study_results} shows the results on CIFAR-10 with $\upgrad$ when varying the batch size from 4 to 64.
Concretely, because we are using SSJD, this makes the number of rows of the sub-Jacobian aggregated at each step vary from 4 to 64.
Recall that IWRM with SSJD and $\meanagg$ is equivalent to ERM with SGD.
We observe that with a small batch size, $\upgrad$ becomes very similar to $\meanagg$.
This is not surprising since both would be equivalent with a batch size of 1.
Conversely, a larger batch size increases the gap between $\upgrad$ and $\meanagg$.
Since the projections of $\upgrad$ are onto the dual cone of more rows, each step becomes non-conflicting with respect to more of the original 1024 objectives, pushing even further the benefits of the non-conflicting property.
In other words, increasing the batch size refines the dual cone, thereby improving the quality of the projections.
\bssresults

\subsection{Compatibility with Adam}\label{appendix:adam_study}

Figure~\ref{fig:adam_study_results} gives the results on CIFAR-10 and SVHN when using \texttt{Adam} rather than the \texttt{SGD} optimizer.
Concretely, this corresponds to the Adam algorithm in which the gradient is replaced by the aggregation of the Jacobian.
The learning rate is still tuned as described in Appendix~\ref{appendix:lr_selection}, but the other hyperparameters of Adam are fixed to the default values of \pytorch, \ie $\beta_1 = 0.9$, $\beta_2 = 0.999$ and $\epsilon = 10^{-8}$.
Because optimization with Adam is faster, the number of epochs for SVHN and CIFAR-10 is reduced to 20 and 15, respectively.
While the performance gap is smaller with this optimizer, it is still significant and suggests that our methods are beneficial with other optimizers than the simple \texttt{SGD}.
The thorough investigation of the interplay between aggregators and momentum-based optimizers is a compelling future research direction.

\adamresults
\clearpage
\section{Applications}\label{appendix:applications}

In this appendix, we discuss briefly about some potential applications for Jacobian descent.

\paragraph{Multi-task learning.}
In multi-task learning, a single model is trained to perform several related tasks simultaneously, leveraging shared representations to improve overall performance~\citep{mtl_survey_ruder}.
At its core, multi-task learning is a multi-objective optimization problem~\citep{mgda2}, making it a straightforward application for Jacobian descent.
Yet, the conflict between tasks is often too limited to justify the overhead of computing all task-specific gradients, \ie the whole Jacobian~\citep{no_moo1, no_moo2}.
In such cases, a practical approach is to minimize some linear scalarization of the objectives using an SGD-based method.
Nevertheless, we believe that a setting with inherent conflict between tasks naturally prescribes Jacobian descent with a non-conflicting aggregator.
We analyze several related works applied to multi-task learning in Section~\ref{sec:related}.

\paragraph{Constrained optimization.}
Constrained optimization problems can be approximated by turning each constraint into an additional objective.
In such a situation, Jacobian descent could be used to respect these extra objectives without having to find weights for them.

\paragraph{Regularization as an additional loss.}
A common technique of regularization is to add an extra term to the loss function.
This term could be considered as a distinct loss, and optimized jointly with the main objective.

\paragraph{Adversarial training.}
In adversarial domain adaptation, the feature extractor of a model is trained with two conflicting objectives:
The features should be helpful for the main task and should be unable to discriminate the domain of the input~\citep{dann}.
Likewise, in adversarial fairness, the feature extractor is trained to both minimize the predictability of sensitive attributes, such as race or gender, and maximize the performance on the main task~\citep{adversarial_fairness}.
Combining the corresponding gradients with a non-conflicting aggregator could improve the optimization efficiency of such methods.
We believe that the training of generative adversarial networks~\citep{gan} could be similarly formulated as a multi-objective optimization problem.
The generator and discriminator could then be jointly optimized with JD.

\paragraph{Momentum-based optimization.}
In gradient-based single-objective optimization, several methods use some form of gradient momentum to improve their convergence speed~\citep{momentum}.
Essentially, their updates consider an exponential moving average of past gradients rather than just the last one.
An appealing idea is to modify those algorithms to make them combine the gradient and the momentum with some aggregator, such as $\upgrad$, instead of summing them.
This would apply to many popular optimizers, like SGD with Nesterov momentum~\citep{nesterov_momentum}, Adam~\citep{adam}, AdamW~\citep{adamw} and NAdam~\citep{nadam}.

\paragraph{Distributed optimization.}
In a distributed data-parallel setting with multiple machines or multiple GPUs, model updates are computed in parallel.
This can be viewed as multi-objective optimization with one objective per data share.
Rather than the typical averaging, a specialized aggregator, such as $\upgrad$, could thus combine the model updates.
This consideration can even be extended to federated learning, in which multiple entities participate in the training of a common model from their own private data by sharing model updates~\citep{fl_survey}.
In this setting, as security is one of the main challenges, the non-conflicting property of the aggregator could be interesting.

\paragraph{Mixed preference optimization.}
When optimizing large language models for human preferences, we may use several loss functions, such as the preference loss, quality loss and generation loss~\citep{mpo}.
These losses could be optimized simultaneously using Jacobian descent, reducing the sensitivity to the weights we give them.

\clearpage
\section{Proofs}\label{appendix:proofs}

\subsection{Preliminary theoretical results}\label{appendix:preliminary}

\begin{lemma}\label{lem:row_span_projection}
Let $\J\in\Rmn$, let $\vy\in\Rn$ and let $\vy'$ be the orthogonal projection of $\vy$ onto the span of the rows of $\J$, then $\J \vy' = \J \vy$ and $\|\vy'\| \leq \|\vy\|$.
\begin{proof}
Let $P\in\mathbb R^{n\times n}$ be the orthogonal projection matrix onto the span of the rows of $\J$, so that $\vy'=P\vy$.
Since every row of $\J$ belongs to this span, $\J=\J P$, and therefore $\J \vy' = \J P \vy = \J \vy$.
Moreover, $I-P$ is the projection matrix onto the orthogonal complement of this span, so by the Pythagorean theorem,
\begin{align*}
\|\vy\|^2 = \|P\vy\|^2 + \big\|(I-P)\vy\big\|^2 \geq \|P\vy\|^2 = \|\vy'\|^2,
\end{align*}
which concludes the proof.
\end{proof}
\end{lemma}

Recall that a function $\vf:\Rn\to\mathbb R^{m}$ is $\pleq$-convex if for all $\vx,\vy\in\mathbb R^{n}$ and any $\lambda\in[0,1]$,
\begin{align*}
\vf\big(\lambda \vx+(1-\lambda)\vy\big)\pleq \lambda \vf(\vx) + (1-\lambda) \vf(\vy).
\end{align*}
\begin{lemma}\label{lem:supporting_hyperplane}
If $\vf:\Rn\to\Rm$ is a continuously differentiable $\pleq$-convex function, then for any pair of vectors $\vx,\vy\in \Rn$, $\jac \vf(\vx) (\vy-\vx) \pleq \vf(\vy)-\vf(\vx)$.
\begin{proof}
\begin{align*}
\jac \vf(\vx) (\vy-\vx) &= \lim_{\lambda\to 0^+} \frac{\vf\big(\vx+\lambda (\vy-\vx)\big)-\vf(\vx)}{\lambda}&\left(\substack{\text{differentiation}}\right)&\\
&\pleq \lim_{\lambda\to 0^+} \frac{\vf(\vx) + \lambda \big(\vf(\vy) - \vf(\vx)\big)-\vf(\vx)}{\lambda}&\left(\substack{\pleq\text{-convexity}}\right)&\\
&=\vf(\vy) - \vf(\vx),
\end{align*}
which concludes the proof.
\end{proof}
\end{lemma}

\begin{lemma}\label{lem:cauchy_shwartz_inequality_operator}
Let $\J\in\Rmn$, let $\vu\in\mathbb R^{m}$ and let $\vx\in\mathbb R^{n}$, then
\begin{align*}
\vu^\T\J \vx\leq \| \vu\|\cdot \| \J \|_\Frob \cdot \| \vx \|
\end{align*}
\begin{proof}
Let $\J_i$ be the $i$th row of $\J$, then
\begin{align*}
\left(\vu^\T\J \vx\right)^2&\leq \| \vu \|^2 \cdot \| \J \vx\|^2&\left(\substack{\text{Cauchy-Schwarz}\\\text{inequality}}\right)&\\
&=\| \vu\|^2\cdot \sum_{i\in\range{m}} \left(\J_i^\T \vx\right)^2\\
&\leq \| \vu\|^2\cdot \sum_{i\in\range{m}} \|\J_i\|^2\cdot \|\vx\|^2&\left(\substack{\text{Cauchy-Schwarz}\\\text{inequality}}\right)&\\
&=\| \vu\|^2\cdot \|\J\|_\Frob^2\cdot \|\vx\|^2,
\end{align*}
which concludes the proof.
\end{proof}
\end{lemma}

Recall that a function $\vf:\Rn\to\mathbb R^{m}$ is $\beta$-smooth if for all $\vx,\vy\in\mathbb R^{n}$, 
\begin{align}
\big\| \jac \vf(\vx)-\jac \vf(\vy)\big\|_\Frob\leq \beta\|\vx-\vy\|\label{eqn:beta_smooth}
\end{align}

\begin{lemma}\label{lem:beta_smoothness_inequality}
Let $\vf:\Rn\to\Rm$ be $\beta$-smooth, then for any $\vw\in\Rm$ and any $\vx,\vy\in\Rn$,
\begin{align}
\vw^\T\big(\vf(\vx)-\vf(\vy)-\jac \vf(\vy)(\vx-\vy)\big)\leq \frac{\beta}{2} \|\vw\|\cdot \|\vx-\vy\|^2\label{eqn:grad_smoothness}
\end{align}
\begin{proof}
\begin{align*}
&\vw^\T\big( \vf(\vx)-\vf(\vy)-\jac \vf(\vy)(\vx-\vy) \big)\\
=~& \vw^\T\left( \int_0^1 \jac \vf\big(\vy+t(\vx-\vy)\big)(\vx-\vy)~dt-\jac \vf(\vy)(\vx-\vy) \right)&\left(\substack{\text{fundamental}\\\text{theorem}\\\text{of calculus}}\right)&\\
=~& \int_0^1  \vw^\T\Big(\jac \vf\big(\vy+t(\vx-\vy)\big) - \jac \vf(\vy) \Big)(\vx-\vy)~dt\\
\leq~& \int_0^1 \|\vw \|\cdot\Big\| \jac \vf\big(\vy+t(\vx-\vy)\big) - \jac \vf(\vy)\Big\|_{\text F}\cdot\|\vx-\vy\|~dt&\left(\substack{\text{Lemma~\ref{lem:cauchy_shwartz_inequality_operator}}}\right)&\\
\leq~& \int_0^1 \|\vw \|\cdot\beta t\cdot\|\vx-\vy\|^2~dt&\left(\substack{\text{$\beta$-smoothness~\ref{eqn:beta_smooth}}}\right)&\\
=~&\frac{\beta}{2} \|\vw\|\cdot\|\vx-\vy\|^2,
\end{align*}
which concludes the proof.
\end{proof}
\end{lemma}

\begin{lemma}\label{lem:projection_positive_homogeneity}
Let $\J\in\Rmn$, let $\vx\in\Rn$ and let $0<a\in\mathbb R$, then $\proj_\J(a\cdot \vx)=a\cdot \proj_\J(\vx)$.
\begin{proof}
Let $C=\{\vy\in\Rn: \0\pleq \J\vy\}$ be the dual cone of the rows of $\J$.
Since $C$ is a closed convex set, the projection onto $C$ is unique, and $\proj_\J(\vx)$ is the point of $C$ closest to $\vx$.
Since $\proj_\J(\vx)\in C$ and $0<a$, we have $\0\pleq a\cdot\J\proj_\J(\vx)=\J\big(a\cdot \proj_\J(\vx)\big)$, so $a\cdot \proj_\J(\vx)\in C$.
Any $\vy\in C$ satisfies $\0\pleq\frac{1}{a}\J\vy=\J\left(\frac{1}{a}\vy\right)$, so $\frac{1}{a}\vy\in C$.
Therefore
\begin{align*}
\big\|a\cdot \proj_\J(\vx)-a\cdot\vx\big\|&=a\big\|\proj_\J(\vx)-\vx\big\|\\
&\leq a\left\|\frac{1}{a}\vy-\vx\right\|&\left(\substack{\text{definition of }\proj_\J(\vx)\\\text{and }\frac{1}{a}\vy\in C}\right)&\\
&=\|\vy-a\cdot\vx\|.
\end{align*}
Hence $a\cdot \proj_\J(\vx)$ is a point of $C$ at least as close to $a\cdot\vx$ as any point of $C$.
By uniqueness of the projection onto $C$, $\proj_\J(a\cdot \vx)=a\cdot \proj_\J(\vx)$.
\end{proof}
\end{lemma}

\subsection{Proposition~\ref{prop:upgrad_duality}}\label{appendix:duality_proposition}

\begin{taggedproposition}{\ref{prop:upgrad_duality}}
Let $\J\in\Rmn$.
For any $\vu\in\Rm$, $\proj_\J(\J^\T \vu)=\J^\T \vw$ with
\begin{align}
\vw&\in\argmin_{\substack{\vv\in\Rm:~\vu\pleq \vv}} \vv^\T \J\J^\T \vv.\tagref{eqn:upgrad_dual}
\end{align}
\begin{proof}
This is a direct consequence of Lemma~\ref{lem:kkt_conditions_proj}.
\end{proof}
\end{taggedproposition}

\begin{lemma}\label{lem:kkt_conditions_proj}
Let $\J\in\Rmn$, $\G=\J\J^\T$, $\vu\in\Rm$.
For any $\vw\in\Rm$ satisfying
\begin{subnumcases}{}
\vu\pleq \vw \label{eqn:upgrad_kkt_dual_feasability}\\
\0\pleq \G \vw \label{eqn:upgrad_kkt_primal_feasability}\\
\vu^\T \G \vw=\vw^\T \G \vw \label{eqn:upgrad_kkt_complementary_slackness}
\end{subnumcases}
we have $\proj_\J(\J^\T \vu) = \J^\T \vw$.
Moreover, $\vw$ satisfies these conditions if and only if
\begin{align*}
\vw&\in\argmin_{\vu\pleq \vv} \vv^\T \G \vv.
\end{align*}
\begin{proof}
Let $\vw\in\Rm$ satisfy (\ref{eqn:upgrad_kkt_dual_feasability}), (\ref{eqn:upgrad_kkt_primal_feasability}) and (\ref{eqn:upgrad_kkt_complementary_slackness}), and let $\vv=\vw-\vu$ and $\vx=\J^\T \vw$.
The projection
\begin{align*}
\proj_\J(\J^\T \vu)= \argmin_{\substack{\vx\in\Rn:\\\0\pleq \J \vx}} \frac{1}{2}\| \vx - \J^\T \vu\|^2
\end{align*}
is a convex program with linear constraints.
Consequently, the KKT conditions are both necessary and sufficient.
The Lagrangian is given by $\mathcal L(\vx, \vv)=\frac{1}{2}\| \vx - \J^\T \vu\|^2-\vv^\T \J \vx$.
The KKT conditions are then given by
\begin{align*}
\begin{cases}
\nabla_\vx \mathcal L(\vx,\vv)=\0\\
\0\pleq \vv\\
\0\pleq \J \vx\\
0=\vv^\T \J \vx
\end{cases}
\end{align*}
We now verify that $\vx$ and $\vv$ satisfy each of these conditions:
\begin{align*}
\nabla_\vx \mathcal L(\vx,\vv)&=\vx-\J^\T \vu-\J^\T \vv=\J^\T \vw-\J^\T \vu-\J^\T (\vw-\vu)=\0\\
\0&\pleq \vw-\vu=\vv&\left(\substack{\text{by (\ref{eqn:upgrad_kkt_dual_feasability})}}\right)&\\
\0&\pleq \G \vw=\J \vx&\left(\substack{\text{by (\ref{eqn:upgrad_kkt_primal_feasability})}}\right)&\\
\vv^\T \J \vx&=(\vw-\vu)^\T \G \vw=\vw^\T \G \vw-\vu^\T \G \vw=0&\left(\substack{\text{by (\ref{eqn:upgrad_kkt_complementary_slackness})}}\right)&
\end{align*}
Since the KKT conditions are sufficient, $\vx=\J^\T \vw$ is the solution of the projection problem, \ie $\proj_\J(\J^\T \vu) = \J^\T \vw$.
This proves the first part.

The program $\min_{\vu\pleq \vv} \frac12 \vv^\T \G \vv$ is also convex with linear constraints, so its KKT conditions are also both necessary and sufficient.
Its Lagrangian is given by $\mathcal L'(\vv, \vmu)=\frac{1}{2}\vv^\T \G \vv-\vmu^\T (\vv-\vu)$, and its KKT conditions are given by
\begin{align*}
&\begin{cases}
\nabla_\vv \mathcal L'(\vv,\vmu)=\0\\
\0\pleq \vmu\\
\vu\pleq \vv\\
0=\vmu^\T (\vv-\vu)
\end{cases}\\
\Leftrightarrow~&\begin{cases}
\vmu=\G \vv\\
\0\pleq \G \vv\\
\vu\pleq \vv\\
\vu^\T \G \vv=\vv^\T \G \vv
\end{cases}
\end{align*}
where the equivalence uses the symmetry of $\G$.
These are exactly (\ref{eqn:upgrad_kkt_dual_feasability}), (\ref{eqn:upgrad_kkt_primal_feasability}) and (\ref{eqn:upgrad_kkt_complementary_slackness}), so $\vw$ satisfies them if and only if it is a minimizer, which concludes the proof.
\end{proof}
\end{lemma}

\subsection{Theorem~\ref{thm:convergence} \label{appendix:convergence_theorem}}

\begin{taggedtheorem}{\ref{thm:convergence}}
Let $\vf:\Rn\to\Rm$ be a $\beta$-smooth and $\pleq$-convex function.
Suppose that the Pareto front $\vf(X^\ast)$ is compact and that for any $\vx\in\Rn$, there is $\vx^\ast\in X^\ast$ satisfying $\vf(\vx^\ast)\pleq \vf(\vx)$.
Let $\vx_1\in\Rn$, and for all $t\in\mathbb N$, $\vx_{t+1}=\vx_t - \eta \upgrad\big(\jac \vf(\vx_t)\big)$, with $\eta=\frac{1}{\beta\sqrt{m}}$.
Let $\vw_t$ be the weights defining $\upgrad\big(\jac \vf(\vx_t)\big)$ as per (\ref{eqn:upgrad_weights}), \ie $\upgrad\big(\jac \vf(\vx_t)\big) = \jac \vf(\vx_t)^\T\cdot \vw_t$.
If $\vw_t$ is bounded, then $\vf(\vx_t)$ converges to $\vf(\vx^\ast)$ for some $\vx^\ast\in X^\ast$.
In other words, $\vf(\vx_t)$ converges to the Pareto front.
\end{taggedtheorem}

To prove the theorem we will need Lemmas~\ref{lem:bound_LR}, \ref{lem:convergence_smoothness} and~\ref{lem:convergence_vanilla} below.

\begin{lemma}\label{lem:bound_LR}
Let $\J\in\Rmn$ and $\vw=\frac{1}{m}\sum_{i=1}^m \vw_i$ be the weights defining $\upgrad(\J)$ as per (\ref{eqn:upgrad_weights}).
Let, as usual, $\G=\J\J^\T$, then, 
\begin{align*}
\vw^\T \G \vw\leq \1^\T \G \vw.
\end{align*}
\begin{proof}
Since $\vv^\T \G \vv = \|\J^\T \vv\|^2$ for any $\vv\in\Rm$, the mapping $\vv\mapsto\vv^\T \G \vv$ is convex.
Therefore
\begin{align*}
&\vw^\T \G \vw\\
\leq~&\frac{1}{m} \sum_i \vw_i^\T \G \vw_i&\left( \substack{\text{Jensen's inequality}} \right)&\\
=~&\frac{1}{m} \sum_i \ve{i}^\T \G \vw_i &\left( \substack{\text{Lemma~\ref{lem:kkt_conditions_proj}, (\ref{eqn:upgrad_kkt_complementary_slackness})}} \right)&\\
\leq~&\frac{1}{m} \sum_{i} \1^\T \G \vw_i &\left( \substack{\text{Lemma~\ref{lem:kkt_conditions_proj}, (\ref{eqn:upgrad_kkt_primal_feasability})}\\\ve{i}\pleq\1} \right)&\\
=~&\1^\T \G \vw,
\end{align*}
which concludes the proof.
\end{proof}
\end{lemma}

\begin{lemma}\label{lem:convergence_smoothness}
Under the assumptions of Theorem~\ref{thm:convergence}, for any $\vw\in\Rm$ and any $t\in\mathbb N$,
\begin{align*}
\vw^\T\big(\vf(\vx_{t+1})-\vf(\vx_t)\big)\leq \frac{\| \vw \|}{\beta \sqrt{m}}\left(\frac{\1}{2\sqrt{m}} - \frac{\vw}{\| \vw \|}\right)^\T \G_t \vw_t.
\end{align*}
\begin{proof}
For all $t\in\mathbb N$, let $\J_t=\jac \vf(\vx_t)$, $\G_t=\J_t \J_t^\T$.
Then $\vx_{t+1}=\vx_t-\eta \upgrad(\J_t) = \vx_t - \eta \J_t^\T \vw_t$.
Therefore
\begin{align*}
&\vw^\T\big(\vf(\vx_{t+1})-\vf(\vx_t)\big)\\
\leq~& -\eta \vw^\T \J_t \J_t^\T \vw_t + \frac{\beta \eta^2}{2} \| \vw \|\cdot \| \J_t^\T \vw_t \|^2&\left(\substack{\text{Lemma~\ref{lem:beta_smoothness_inequality}}}\right)&\\
=~& -\frac{1}{\beta \sqrt{m}}\vw^\T \G_t \vw_t + \frac{1}{2 \beta m} \| \vw \|\cdot \vw_t^\T \G_t \vw_t&\left(\substack{\eta=\frac{1}{\beta\sqrt{m}}}\right)&\\
\leq~& -\frac{1}{\beta \sqrt{m}}\vw^\T \G_t \vw_t + \frac{1}{2 \beta m} \| \vw \|\cdot \1^\T \G_t \vw_t&\left( \substack{\text{Lemma~\ref{lem:bound_LR}}} \right)&\\
=~& \frac{\| \vw \|}{\beta \sqrt{m}}\left(\frac{\1}{2\sqrt{m}} - \frac{\vw}{\| \vw \|}\right)^\T \G_t \vw_t,
\end{align*}
which concludes the proof.
\end{proof}
\end{lemma}

\begin{lemma}\label{lem:convergence_vanilla}
Under the assumptions of Theorem~\ref{thm:convergence}, if $\vx^\ast\in X^\ast$ satisfies $\1^\T \vf(\vx^\ast)\leq \1^\T \vf(\vx_t)$ for all $t\in\mathbb N$, then
\begin{align}
\frac{1}{T}\sum_{t\in\range{T}} \vw_t^\T \big(\vf(\vx_t) - \vf(\vx^\ast)\big) \leq \frac{1}{T}\bigg( \1^\T\big(\vf(\vx_1)-\vf(\vx^\ast)\big) + \frac{\beta \sqrt{m}}{2}\left\| \vx_1-\vx^\ast \right\|^2 \bigg).\label{eqn:convergence_bound3}
\end{align}
\begin{proof}
We first bound, for any $t\in\mathbb N$, $\1^\T \big(\vf(\vx_{t+1})-\vf(\vx_t)\big)$ as follows
\begin{align*}
&\1^\T \big(\vf(\vx_{t+1})-\vf(\vx_t)\big)\\
\leq~& -\frac{1}{2\beta \sqrt{m}}\cdot \1^\T \G_t \vw_t&\left( \substack{\text{Lemma~\ref{lem:convergence_smoothness}}\\\text{with }\vw=\1} \right)&\\
\leq~& -\frac{1}{2\beta \sqrt{m}}\cdot \vw_t^\T \G_t \vw_t.&\left( \substack{\text{Lemma~\ref{lem:bound_LR}}} \right)&
\end{align*}
Summing this over $t\in\range{T}$ yields
\begin{align}
&\frac{1}{2\beta \sqrt{m}} \sum_{t\in\range{T}} \vw_t^\T \G_t \vw_t\nonumber\\
\leq~& \sum_{t\in\range{T}} \1^\T \big(\vf(\vx_t)-\vf(\vx_{t+1})\big)\nonumber\\
=~& \1^\T\big(\vf(\vx_1)-\vf(\vx_{T+1})\big)&\left( \substack{\text{Telescoping sum}} \right)&\nonumber\\
\leq~&  \1^\T\big(\vf(\vx_1)-\vf(\vx^\ast)\big).&\left( \substack{\text{Assumption}\\\1^\T \vf(\vx^\ast)\leq \1^\T \vf(\vx_{T+1})} \right)&\label{eqn:convergence_bound2}
\end{align}

Since $\0 \pleq \vw_t$,
\begin{align*}
&\vw_t^\T \big(\vf(\vx_t) - \vf(\vx^\ast)\big)\nonumber\\
\leq~& \vw_t^\T \J_t (\vx_t - \vx^\ast)&\left( \substack{\text{Lemma~\ref{lem:supporting_hyperplane}}} \right)&\\
=~&\frac{1}{\eta} (\vx_t-\vx_{t+1})^\T (\vx_t-\vx^\ast)&\left( \substack{\vx_{t+1}=\vx_t-\eta \J_t^\T \vw_t} \right)&\\
=~&\frac{1}{2\eta} \Big( \| \vx_t -\vx_{t+1} \|^2 + \| \vx_t-\vx^\ast \|^2 - \| \vx_{t+1} - \vx^\ast \|^2  \Big)&\left( \substack{\text{Parallelogram}\\\text{law}} \right)&\\
=~&\frac{1}{2\beta \sqrt{m}} \vw_t^\T \G_t \vw_t + \frac{\beta \sqrt{m}}{2}\Big(\| \vx_t-\vx^\ast \|^2 - \| \vx_{t+1} - \vx^\ast \|^2  \Big).&\left( \substack{\eta=\frac{1}{\beta\sqrt{m}}} \right)&
\end{align*}
Summing this over $t\in\range{T}$ yields
\begin{align*}
&\sum_{t\in\range{T}} \vw_t^\T \big(\vf(\vx_t) - \vf(\vx^\ast)\big)\\
\leq~& \frac{1}{2\beta \sqrt{m}} \sum_{t\in\range{T}} \vw_t^\T \G_t \vw_t + \frac{\beta \sqrt{m}}{2}\Big(\| \vx_1-\vx^\ast \|^2 - \| \vx_{T+1} - \vx^\ast \|^2\Big)&\left( \substack{\text{Telescoping sum}} \right)&\\
\leq~&\frac{1}{2\beta \sqrt{m}} \sum_{t\in\range{T}} \vw_t^\T \G_t \vw_t + \frac{\beta \sqrt{m}}{2}\| \vx_1-\vx^\ast \|^2\\
\leq~&\1^\T\big(\vf(\vx_1)-\vf(\vx^\ast)\big) + \frac{\beta \sqrt{m}}{2}\| \vx_1-\vx^\ast \|^2.&\left( \substack{\text{By (\ref{eqn:convergence_bound2})}} \right)&
\end{align*}
Scaling down this inequality by $T$ yields
\begin{align*}
\frac{1}{T}\sum_{t\in\range{T}} \vw_t^\T \big(\vf(\vx_t) - \vf(\vx^\ast)\big) \leq \frac{1}{T}\bigg( \1^\T(\vf(\vx_1)-\vf(\vx^\ast)) + \frac{\beta \sqrt{m}}{2}\| \vx_1-\vx^\ast \|^2 \bigg),
\end{align*}
which concludes the proof.
\end{proof}
\end{lemma}

We are now ready to prove Theorem~\ref{thm:convergence}.

\begin{proof}
For all $t\in\mathbb N$, let $\J_t=\jac \vf(\vx_t)$, $\G_t=\J_t \J_t^\T$.
Then 
\begin{align*}
\vx_{t+1}&=\vx_t-\eta \upgrad(\J_t)\\
&= \vx_t - \eta\J_t^\T \vw_t.
\end{align*}

Substituting $\vw=\1$ in the term $\frac{\1}{2\sqrt{m}} - \frac{\vw}{\| \vw \|}$ of Lemma~\ref{lem:convergence_smoothness} yields 
\begin{align*}
\frac{\1}{2\sqrt{m}} - \frac{\vw}{\| \vw \|}&=-\frac{\1}{2\sqrt{m}}\\
&\plt\0.
\end{align*}
Therefore there exists some $\varepsilon > 0$ such that any $\vw\in\Rm$ with $\| \1 - \vw\|<\varepsilon$ satisfies $\frac{\1}{2\sqrt{m}} \plt \frac{\vw}{\| \vw \|}$.
Denote by $B_\varepsilon(\1)=\{ \vw\in\Rm : \| \1 - \vw \| < \varepsilon \}$, \ie for all $\vw\in B_\varepsilon(\1)$, $\frac{\1}{2\sqrt{m}} \plt \frac{\vw}{\| \vw \|}$.
By the non-conflicting property of $\upgrad$, $\0\pleq \G_t \vw_t$ and therefore for all $\vw\in B_\varepsilon(\1)$, 
\begin{align*}
&\vw^\T\big(\vf(\vx_{t+1})-\vf(\vx_t)\big)\\
\leq~& \frac{\| \vw\|}{\beta\sqrt{m}}\left( \frac{\1}{2\sqrt{m}} - \frac{\vw}{\| \vw \|} \right) \G_t \vw_t&\left( \substack{\text{Lemma~\ref{lem:convergence_smoothness}}} \right)&\\
\leq~& 0.
\end{align*}
Since $\vw^\T \vf(\vx_t)$ is bounded and non-increasing, it converges.
Since $B_\varepsilon(\1)$ contains a basis of $\Rm$, $\vf(\vx_t)$ converges to some $\vf^\ast\in\Rm$.
By the assumption on $\vf$, for each $t\in\mathbb N$, there exists $\vx_t^\ast$ in the Pareto set such that $\vf(\vx_t^\ast)\pleq\vf(\vx_t)$.
Given that the Pareto front is compact, the sequence $\{\vf(\vx_t^\ast)\}_{t\in\mathbb N}$ admits a subsequence converging to some $\vf(\vx^\ast)$ also in the Pareto front.
Since $\vf(\vx_t)\to\vf^\ast$, the closedness of the partial order $\pleq$ implies that $\vf(\vx^\ast)\pleq\vf^\ast$.

We now prove that $\vf(\vx^\ast)= \vf^\ast$.
Since $\vf(\vx^\ast)\pleq \vf^\ast$, it is sufficient to show that $\1^\T \big(\vf^\ast-\vf(\vx^\ast)\big)\leq 0$.

First, the additional assumption of Lemma~\ref{lem:convergence_vanilla} applies since $\1^\T \vf(\vx_t)$ decreases to $\1^\T \vf^\ast$ which is larger than or equal to $\1^\T \vf(\vx^\ast)$.
Therefore
\begin{align}
&\1^\T \big(\vf^\ast-\vf(\vx^\ast)\big)\nonumber\\
\leq~& \left( \frac{m}{T} \sum_{t\in\range{T}} \vw_t \right)^\T \big(\vf^\ast-\vf(\vx^\ast)\big)&\left( \substack{\vf(\vx^\ast)\pleq \vf^\ast\\\1\pleq m\vw_t\\\text{ by (\ref{eqn:upgrad_kkt_dual_feasability})}} \right)&\nonumber\\
=~&\frac{m}{T} \sum_{t\in\range{T}} \vw_t^\T\Big(\vf^\ast - \vf(\vx_t) + \vf(\vx_t) - \vf(\vx^\ast) \Big)\nonumber\\
=~&\frac{m}{T} \left(\sum_{t\in\range{T}} \vw_t^\T\big(\vf^\ast - \vf(\vx_t)\big) + \sum_{t\in\range{T}} \vw_t^\T\big(\vf(\vx_t) - \vf(\vx^\ast)\big) \right)\nonumber\\
\leq~&\frac{m}{T} \left(\sum_{t\in\range{T}} \vw_t^\T\big(\vf^\ast - \vf(\vx_t)\big) + \1^\T\big(\vf(\vx_1)-\vf(\vx^\ast)\big) + \frac{\beta \sqrt{m}}{2}\| \vx_1-\vx^\ast \|^2 \right)&\left( \substack{\text{Lemma~\ref{lem:convergence_vanilla}}} \right)&\label{eqn:pre_convergence_rate}
\end{align}

Taking the limit as $T\to\infty$, we get
\begin{align*}
&\1^\T \big(\vf^\ast-\vf(\vx^\ast)\big)\\
\leq~&\lim_{T\to\infty}\frac{m}{T}\sum_{t\in\range{T}} \vw_t^\T\big(\vf^\ast - \vf(\vx_t)\big)\\
\leq~& \lim_{T\to\infty}\frac{m}{T}\sum_{t\in\range{T}} \|\vw_t\|\cdot\big\|\vf^\ast-\vf(\vx_t)\big\|&\left( \substack{\text{Cauchy-Schwarz}\\\text{inequality}} \right)&\\
=~&0,&\left( \substack{\vw_t\text{ bounded}\\\vf(\vx_t) \to \vf^\ast} \right)&
\end{align*}
which concludes the proof.
\end{proof}

\subsection{Supplementary theoretical results \label{appendix:supplementary_theory}}

The proof of Theorem~\ref{thm:convergence} provides some additional insights about the convergence rate as well as some notion of convergence in the non-convex case.

\paragraph{Convergence rate.}
Combining the equality $\vf^\ast=\vf(\vx^\ast)$ and (\ref{eqn:pre_convergence_rate}), we have
\begin{align*}
\frac{1}{T} \sum_{t\in\range{T}} \vw_t^\T\big(\vf(\vx_t) - \vf(\vx^\ast)\big) \leq \frac{1}{T}\1^\T\big(\vf(\vx_1)-\vf(\vx^\ast)\big) + \frac{\beta \sqrt{m}}{2T}\| \vx_1-\vx^\ast \|^2
\end{align*}
Furthermore, the Cauchy-Schwarz inequality yields $\1^\T\big(\vf(\vx_1)-\vf(\vx^\ast)\big) \leq \sqrt{m}\cdot\big\| \vf(\vx_1)-\vf(\vx^\ast) \big\|$, so
\begin{align}
\frac{1}{T} \sum_{t\in\range{T}} \vw_t^\T\big(\vf(\vx_t) - \vf(\vx^\ast)\big) \leq \frac{\sqrt{m}}{T}\bigg(\big\| \vf(\vx_1)-\vf(\vx^\ast) \big\| + \frac{\beta}{2}\| \vx_1-\vx^\ast \|^2\bigg)\label{eqn:convergence_rate}
\end{align}
This hints a convergence rate of order $\bigo\left( \frac{1}{T} \right)$, whose constant depends on the initial point $\vx_1$.

\paragraph{Non-convex setting.} The proof of (\ref{eqn:convergence_bound2}) does not require the convexity of the objective function. This bound can be equivalently formulated as
\begin{align}
\frac{1}{T}\sum_{t=1}^T \big\| J_t^\top w_t \big\|^2 \leq \frac{2\beta\sqrt{m}}{T}\mathbf 1^\top \big( f(x_1)-f(x^\ast) \big)
\end{align}
This shows the convergence of the updates in the non-convex setting, under the smoothness condition of Theorem~\ref{thm:convergence}.

\subsection{Derivations of Section~\ref{sec:implementation}}\label{appendix:gramian_based_derivations}

In this section, we detail the derivations of (\ref{eqn:taylor-approxmiation-gramian}) and (\ref{eqn:gramian_based_update}).
Let $\vf:\Rn\to\Rm$ be continuously differentiable, let $\vx\in\Rn$, and let $\A$ be a weighted aggregator.

\paragraph{Taylor approximation in terms of the Gramian.}
For a fixed $\vx\in\Rn$, let $\vw\in\Rm$ be such that $\A\big(\jac \vf(\vx)\big)=\jac \vf(\vx)^\T \vw$, and let $\vy=-\eta\jac \vf(\vx)^\T \vw$ be the corresponding update of JD.
Observe that
\begin{align*}
\|\vy\|^2 = \eta^2 \vw^\T \jac \vf(\vx) \jac \vf(\vx)^\T \vw = \eta^2 \vw^\T \gram \vf(\vx) \vw.
\end{align*}
Substituting $\vy$ into the Taylor approximation (\ref{eqn:taylor-approximation}) thus yields
\begin{align*}
\vf(\vx+\vy)&=\vf(\vx)+\jac \vf(\vx)\cdot \vy+o(\|\vy\|)\\
&=\vf(\vx)-\eta\jac \vf(\vx) \jac \vf(\vx)^\T \vw+o\left(\eta\big\|\jac \vf(\vx)^\T \vw\big\|\right)\\
&=\vf(\vx)-\eta \gram \vf(\vx)\cdot \vw + o\left(\eta\sqrt{\vw^\T\cdot \gram \vf(\vx) \cdot \vw}\right),
\end{align*}
which is (\ref{eqn:taylor-approxmiation-gramian}).

\paragraph{Gramian-based update.}
Suppose now that $\A(\J)=\J^\T\cdot \W(\G)$ for some $\W:\PSD{m}\to\Rm$, where $\G=\J\J^\T$.
For any constant vector $\vw\in\Rm$, the function $\vw^\T\vf:\Rn\to\R$ is a linear combination of the coordinates of $\vf$, so by linearity of differentiation,
\begin{align*}
\nabla\left(\vw^\T\vf\right)(\vx)=\sum_{i\in\range{m}} w_i \nabla f_i(\vx)=\jac \vf(\vx)^\T \vw.
\end{align*}
Let $\vw=\W\big(\gram \vf(\vx)\big)$.
Note that $\vw$ depends on $\vx$, but it is computed before the differentiation step and then treated as a constant, \ie we do not differentiate through $\W$ nor through $\gram \vf$.
Therefore,
\begin{align*}
\A\big(\jac \vf(\vx)\big)&=\jac \vf(\vx)^\T\cdot \W\big(\gram \vf(\vx)\big)\\
&=\jac \vf(\vx)^\T \vw\\
&=\nabla\left(\vw^\T\vf\right)(\vx)\\
&=\nabla \Big(\W\big(\gram \vf(\vx)\big)^\T \cdot \vf\Big)(\vx),
\end{align*}
which is (\ref{eqn:gramian_based_update}).

\clearpage
\section{Properties of existing aggregators}\label{appendix:properties}
In the following, we prove the properties of the aggregators from Table~\ref{tab:properties}.
Some aggregators, \eg $\rgw$, $\graddrop$ and $\pcgrad$, are non-deterministic and are thus not technically functions but rather random variables whose distribution depends on the matrix $\J \in \Rmn$ to aggregate. Still, the properties of Section~\ref{subsection:properties} can be easily adapted to a random setting.
If $\A$ is a random aggregator, then for any $\J$, $\A(\J)$ is a random vector in $\Rn$. 
The aggregator is non-conflicting if $\A(\J)$ is in the dual cone of the rows of $\J$ with probability 1.
It is linear under scaling if for all $\J \in \Rmn$, there is a -- possibly random -- matrix $\mathfrak{\J} \in \Rmn$ such that for all $\0 \plt \vc \in \Rm$, $\A(\diag(\vc) \cdot \J) = \mathfrak{\J}^\T \cdot \vc$.
Finally, $\A$ is weighted if for any $\J\in\Rmn$ there is a -- possibly random -- weighting $\vw\in\Rm$ satisfying $\A(\J)=\J^\T\cdot \vw$.

\subsection{Mean}
$\meanagg$ simply averages the rows of the input matrix, \ie for all $\J \in \Rmn$,
\begin{equation}
\meanagg(\J) = \frac{1}{m}\J^\T \cdot \1\label{eqn:mean}
\end{equation}

\paragraph{\xmark\, Non-conflicting.}
$\meanagg\left(\begin{bmatrix} -2\\4 \end{bmatrix}\right) = \begin{bmatrix}1\end{bmatrix}$, which conflicts with $\begin{bmatrix}-2\end{bmatrix}$, so $\meanagg$ is not non-conflicting.

\paragraph{\cmark\, Linear under scaling.}
For any $\vc \in \Rm$, $\meanagg(\diag(\vc) \cdot \J) = \frac{1}{m}\J^\T \cdot \vc$, which is linear in $\vc$.
$\meanagg$ is therefore linear under scaling.

\paragraph{\cmark\, Weighted.}
By (\ref{eqn:mean}), $\meanagg$ is weighted with constant weighting equal to $\frac{1}{m}\1$.

\subsection{MGDA}
The optimization algorithm presented in~\citet{mgda1}, called MGDA, is tied to a particular method for aggregating the gradients.
We thus refer to this aggregator as $\mgda$.
The dual problem of this method was also introduced independently in~\citet{fs}.
We show the equivalence between the two solutions to make the analysis of $\mgda$ easier.

For all $\J\in\Rmn$, the aggregation described in~\citet{mgda1} is defined as
\begin{align}
\mgda(\J) &= \J^\T \cdot \vw\label{eqn:mgda}\\
\text{with}\hspace{0.45cm}\vw &\in \argmin_{\substack{\0\pleq \vv:\\\1^\T \vv=1}} \left\| \J^\T \vv\right\|^2\label{eqn:mgda_weights}
\end{align}

In Equation (3) of~\citet{fs}, the following problem is studied:
\begin{align}
\min_{\substack{\alpha\in\mathbb R, \vx\in\Rn:\\\J \vx\pleq \alpha \1}} \alpha + \frac12 \| \vx\|^2\label{eqn:fs}
\end{align}

We show that the problems in (\ref{eqn:mgda_weights}) and (\ref{eqn:fs}) are dual to each other.
Furthermore, the duality gap is null since this is a convex problem.
The Lagrangian of the problem in (\ref{eqn:fs}) is given by $\mathcal L(\alpha, \vx,\vmu)=\alpha + \frac12 \| \vx\|^2 - \vmu^\T(\alpha \1 - \J \vx)$.
Differentiating \wrt $\alpha$ and $\vx$ gives respectively $1- \1^\T \vmu$ and $\vx+\J^\T \vmu$.
The dual problem is obtained by setting those two to $\0$ and then maximizing the Lagrangian on $\0\pleq \vmu$ and $\alpha$, \ie
\begin{align*}
&\argmax_{\substack{\alpha, \0\pleq \vmu:\\\1^\T \vmu=1}} \alpha + \frac12 \left\| \J^\T \vmu\right\|^2 - \vmu^\T\left(\alpha \1 + \J \J^\T \vmu\right)\\
=~&\argmax_{\substack{\alpha, \0\pleq \vmu:\\\1^\T \vmu=1}} \alpha + \frac12 \left\| \J^\T \vmu\right\|^2 - \alpha \vmu^\T\1 - \vmu^\T\J \J^\T \vmu\\
=~&\argmin_{\substack{\0\pleq \vmu:\\\1^\T \vmu=1}} \frac12 \left\| \J^\T \vmu\right\|^2
\end{align*}
Therefore, (\ref{eqn:mgda_weights}) and (\ref{eqn:fs}) are equivalent, with $\vx = - \J^\T \vw$.

\paragraph{\cmark\, Non-conflicting.}
Observe that since in (\ref{eqn:fs}), $\alpha=0$ and $\vx=\0$ is feasible, the objective is non-positive and therefore $\alpha\leq 0$.
Substituting $\vx = -\J^\T \vw$ in $\J \cdot \vx\pleq \alpha \1 \pleq \0$ yields $\0\pleq \J \J^\T \vw$, \ie $\0\pleq \J \cdot \mgda(\J)$, so $\mgda$ is non-conflicting.

\paragraph{\xmark\, Linear under scaling.}
With $\J = \begin{bmatrix} 2&0\\0&2\\a&a \end{bmatrix}$, if $0 \leq a \leq 1$, $\mgda(\J) = \begin{bmatrix} a \\ a\end{bmatrix}$.
However, if $a \geq 1$, $\mgda(\J) = \begin{bmatrix} 1 \\ 1\end{bmatrix}$.
This is not affine in $a$, so $\mgda$ is not linear under scaling. In particular, if any row of $\J$ is $\0$, $\mgda(\J) = \0$.
This implies that the optimization will stop whenever one objective has converged.

\paragraph{\cmark\, Weighted.}
By (\ref{eqn:mgda}), $\mgda$ is weighted.

\subsection{DualProj}
The projection of a gradient of interest onto a dual cone was first described in~\citet{gem}.
When this gradient is the average of the rows of the Jacobian, we call this aggregator $\dualproj$.
Formally,
\begin{equation}
\dualproj(\J) = \frac{1}{m}\cdot \proj_\J\left(\J^\T \cdot \1\right)
\end{equation}
where $\proj_\J$ is the projection operator defined in (\ref{eqn:projection}).

\paragraph{\cmark\, Non-conflicting.}
By the constraint in (\ref{eqn:projection}), $\dualproj$ is non-conflicting.

\paragraph{\xmark\, Linear under scaling.}
With $\J = \begin{bmatrix} 2&0\\-2a&2a \end{bmatrix}$, if $a \geq 1$, $\dualproj(\J) = \begin{bmatrix} 0\\a \end{bmatrix}$.
However, if $0.5 \leq a \leq 1$, $\dualproj(\J) = \begin{bmatrix} 1-a\\a \end{bmatrix}$.
This is not affine in $a$ on $[0.5, +\infty[$, so $\dualproj$ is not linear under scaling.

\paragraph{\cmark\, Weighted.}
By Proposition~\ref{prop:upgrad_duality}, $\dualproj(\J) = \frac{1}{m}\J^\T \cdot \vw$, with $\vw\in\argmin_{\1\pleq \vv} \vv^\T \J\J^\T \vv$.
$\dualproj$ is thus weighted.

\subsection{PCGrad}

$\pcgrad$ is described in~\citet{pcgrad}.
It projects each gradient onto the orthogonal hyperplane of other gradients in case of conflict with them, iteratively and in random order.
When $m \leq 2$, $\pcgrad$ is deterministic and satisfies $\pcgrad = m \cdot \upgrad$.
Therefore, in this case, it satisfies all three properties.
When $m > 2$, $\pcgrad$ is non-deterministic, so $\pcgrad(\J)$ is a random vector.

For any index $i \in \range{m}$, let $\vg_i = \J^\T \cdot \ve{i}$ and let $\rvp(i)$ be a random vector distributed uniformly on the set of permutations of the elements in $\range{m}\setminus\{i\}$. For instance, if $m=3$, $\rvp(2) = \begin{bmatrix}1 & 3\end{bmatrix}^\T$ with probability $0.5$ and $\rvp(2) = \begin{bmatrix}3 & 1\end{bmatrix}^\T$ with probability $0.5$.
For notation convenience, whenever $i$ is clear from context, we denote $j_k = \rvp(i)_k$.
The iterative projection of $\pcgrad$ is then defined recursively as:
\begin{align}
\vg_{i,1}^{\text{PC}} &= \vg_i\\
\vg_{i,k+1}^{\text{PC}} &= \vg_{i,k}^{\text{PC}} - \mathds{1}\{\vg_{i,k}^{\text{PC}} \cdot \vg_{j_k} < 0\} \frac{\vg_{i,k}^{\text{PC}} \cdot \vg_{j_k}}{\|\vg_{j_k}\|^2}\vg_{j_k}\label{eqn:pcgrad_projection}
\end{align}

We noticed that an equivalent formulation to the conditional projection of (\ref{eqn:pcgrad_projection}) is the projection onto the dual cone of $\{\vg_{j_k}\}$:
\begin{align}
\vg_{i,k+1}^{\text{PC}} = \proj_{\vg_{j_k}^\T}(\vg_{i,k}^{\text{PC}}) \label{eqn:pcgrad_dual_cone_projection}
\end{align}

Finally, the aggregation is given by
\begin{equation}
\pcgrad(\J) = \sum_{i=1}^m \vg_{i,m}^{\text{PC}}.
\end{equation}

\paragraph{\xmark\, Non-conflicting.}
If $\J=\begin{bmatrix} 1&0\\0&1\\-0.5&-1 \end{bmatrix}$, the only non-conflicting direction is $\0$. However, $\pcgrad(\J)$ is uniform over the set $\left\{ \begin{bmatrix}0.4 \\ 0.2\end{bmatrix}, \begin{bmatrix}0.8 \\ 0.2\end{bmatrix}, \begin{bmatrix}0.4 \\ -0.2\end{bmatrix}, \begin{bmatrix}0.8 \\ -0.2\end{bmatrix} \right\}$, \ie $\pcgrad(\J)$ is in the dual cone of the rows of $\J$ with probability $0$.
$\pcgrad$ is thus not non-conflicting.
Here, $\mathds{E}[\pcgrad(\J)] = \begin{bmatrix}0.6 & 0\end{bmatrix}^\T$, so $\pcgrad$ is neither non-conflicting in expectation.

\paragraph{\cmark\, Linear under scaling.}
To show that $\pcgrad$ is linear under scaling, let $\0 \plt \vc \in \Rm$, $\vg_i' = c_i \vg_i$, $\vg_{i,1}^{\prime\text{ PC}} = \vg'_i$ and $\vg_{i, k+1}^{\prime\text{ PC}} = \proj_{\vg_{j_k}^{\prime \T}}\big(\vg_{i,k}^{\prime\text{ PC}}\big)$.
We show by induction that $\vg_{i, k}^{\prime\text{ PC}} = c_i \vg_{i, k}^{\text{PC}}$.

The base case is given by $\vg_{i,1}^{\prime\text{ PC}} = \vg_i' = c_i \vg_i = c_i \vg_{i,1
}^{\text{PC}}$.

Then, assuming the induction hypothesis $\vg_{i, k}^{\prime\text{ PC}} = c_i \vg_{i, k}^{\text{PC}}$, we show $\vg_{i, k+1}^{\prime\text{ PC}} = c_i \vg_{i, k+1}^{\text{PC}}$:

\begin{align*}
&\vg_{i, k+1}^{\prime\text{ PC}} = \proj_{c_{j_k}\vg_{j_k}^\T}\big(c_i \vg_{i,k}^{\text{ PC}}\big)&\left( \substack{\text{Induction hypothesis}} \right)\\
&\vg_{i, k+1}^{\prime\text{ PC}} = c_i \proj_{\vg_{j_k}^\T}\big(\vg_{i,k}^{\text{ PC}}\big) &\left( \substack{0 < c_i \text{ and } 0 < c_{j_k}} \right)&\\
&\vg_{i, k+1}^{\prime\text{ PC}} = c_i \vg_{i, k+1}^{\text{PC}}&\left( \substack{\text{By (\ref{eqn:pcgrad_dual_cone_projection})}} \right)&
\end{align*}

Therefore $\pcgrad\big(\diag(\vc) \cdot \J\big) = \sum_{i=1}^m c_i \vg_{i,m}^{\text{PC}}$, so it can be written as $\pcgrad\big(\diag(\vc) \cdot \J\big) = \mathfrak{\J}^\T\cdot \vc$ with $\mathfrak{\J}=\begin{bmatrix}\vg_{1,m}^{\text{PC}}&\cdots&\vg_{m,m}^{\text{PC}}\end{bmatrix}^\T$.
Therefore, $\pcgrad$ is linear under scaling.

\paragraph{\cmark\, Weighted.}
For all $i$, $\vg_{i,m}^{\text{PC}}$ is always a random linear combination of rows of $\J$.
$\pcgrad$ is thus weighted.

\subsection{GradDrop}
The aggregator used by the GradDrop layer, which we denote $\graddrop$, is described in~\citet{graddrop}.
It is non-deterministic, so $\graddrop(\J)$ is a random vector.
Given $\J\in\Rmn$, let $|\J| \in \Rmn$ be the element-wise absolute value of $\J$.
Let $P=\frac{1}{2}\left(\1 + \frac{\J^\T\cdot \1}{|\J|^\T \cdot\1}\right) \in \Rn$, where the division is element-wise.
Each coordinate $i\in [n]$ is independently assigned to the set $\mathcal I_+$ with probability $P_i$ and to the set $\mathcal I_-$ otherwise.
The aggregation at coordinate $i\in \mathcal I_+$ is given by the sum of all positive $\J_{ji}$, for $j\in \range{m}$.
The aggregation at coordinate $i\in \mathcal I_-$ is given by the sum of all negative $\J_{ji}$, for $j\in \range{m}$.
Formally,
\begin{align}
\graddrop(\J) &= \left(\sum_{i\in\mathcal I_+} \ve{i}\sum_{\substack{j\in\range{m}:\\\J_{ji}>0}}\J_{ji}\right) + 
\left(\sum_{i\in\mathcal I_-}\ve{i}\sum_{\substack{j\in\range{m}:\\\J_{ji}<0}}\J_{ji}\right)
\end{align}

\paragraph{\xmark\, Non-conflicting.}
If $\J=\begin{bmatrix}-2\\1\end{bmatrix}$, then $P=\begin{bmatrix} 1/3 \end{bmatrix}$.
Therefore, $\mathds{P}\big[\graddrop(\J) = \begin{bmatrix}-2\end{bmatrix}\big] = 2/3$ and $\mathds{P}\big[\graddrop(\J) = \begin{bmatrix}1\end{bmatrix}\big] = 1/3$, \ie $\graddrop(\J)$ is in the dual cone of the rows of $\J$ with probability $0$.
Therefore, $\graddrop$ is not non-conflicting.
Here, $\mathds{E}\big[\graddrop(\J)\big] = \begin{bmatrix}-1\end{bmatrix}^\T$, so $\graddrop$ is neither non-conflicting in expectation.

\paragraph{\xmark\, Linear under scaling.}
If $\J=\begin{bmatrix}1&-1\\-1&1\end{bmatrix}$, then $P=\frac{1}{2}\cdot \1$ and the aggregation is one of the four vectors $\begin{bmatrix} \pm1&\pm1 \end{bmatrix}^\T$ with equal probability.
Scaling the first line of $\J$ by $2$ yields $\J=\begin{bmatrix}2&-2\\-1&1\end{bmatrix}$ and $P=\begin{bmatrix}2/3 & 1/3\end{bmatrix}^\T$, which cannot lead to a uniform distribution over four elements. Therefore, $\graddrop$ is not linear under scaling.

\paragraph{\xmark\, Weighted.}
With $\J=\begin{bmatrix}1&-1\\-1&1\end{bmatrix}$, the span of $\J$ does not include $\begin{bmatrix}1&1\end{bmatrix}^\T$ nor $\begin{bmatrix}-1&-1\end{bmatrix}^\T$.
Therefore, $\graddrop$ is not weighted.

\subsection{IMTL-G}\label{appendix:imtlg}
In~\citet{imtl}, the authors describe a method to impartially balance gradients by weighting them.
Let $\vg_i$ be the $i$'th row of $\J$ and let $\vu_i = \frac{\vg_i}{\|\vg_i\|}$.
They want to find a combination $\vg=\sum_{i=1}^m \alpha_i \vg_i$ such that $\vg^\T \vu_i$ is equal for all $i$.
Let $U=\begin{bmatrix} \vu_1-\vu_2&\dots&\vu_1-\vu_m\end{bmatrix}^\T$, $D=\begin{bmatrix} \vg_1-\vg_2&\dots&\vg_1-\vg_m \end{bmatrix}^\T$.
If $\valpha_{2:m}=\begin{bmatrix} \alpha_2 & \dots&\alpha_m \end{bmatrix}^\T$, then $\valpha_{2:m}=\big(UD^\T\big)^{-1} U \cdot \vg_1$ and $\alpha_1=1-\sum_{i=2}^m \alpha_i$.
Notice that this is defined only when the gradients are linearly independent.
Thus, this is not strictly speaking an aggregator since it can only be computed on matrices of rank $m$.
We thus propose a generalization defined for matrices of any rank that is equivalent when the matrix has rank $m$.
In the original formulation, requiring $\vg^\T \vu_i$ to be equal to some $c\in\R$ for all $i$, is equivalent to requiring that for all $i$, $\vg^\T \vg_i$ is equal to $c~\| \vg_i\|$.
Writing $\vg=\J^\T \valpha$ and letting $\vd\in\Rm$ be the vector of norms of the rows of $\J$, the objective is thus to find $\valpha$ satisfying $\J \J^\T \valpha \propto \vd$.
Besides, to match the original formulation, the elements of $\valpha$ should sum to $1$.

Letting $\big(\J\J^\T\big)^\dagger$ be the Moore-Penrose pseudo inverse  of $\J\J^\T$, we define
\begin{align}
\imtlg(\J) &= \J^\T \cdot \vw\label{eqn:imtlg}\\
\text{with}\hspace{0.70cm}\vw &= \begin{cases}
    \frac{\vv}{\1^\T\vv},&\text{if }\1^\T\vv \neq \0\\
    \0,&\text{otherwise}
\end{cases}\\
\text{and}\hspace{0.88cm}\vv &= \big(\J \J^\T\big)^\dagger \cdot \vd.
\end{align}

\paragraph{\xmark\, Non-conflicting.}
If $\J=\begin{bmatrix} 1&-1&-1 \end{bmatrix}^\T$, then $\vd=\begin{bmatrix} 1&1&1 \end{bmatrix}^\T$, $\J\J^\T=\begin{bmatrix} 1&-1&-1\\-1&1&1\\-1&1&1 \end{bmatrix}$, and thus $\vv=\frac{1}{9}\begin{bmatrix} -1 & 1 & 1\end{bmatrix}^\T$.
Therefore, $\vw=\begin{bmatrix} -1&1&1 \end{bmatrix}^\T$, $\imtlg(\J) = \begin{bmatrix} -3 \end{bmatrix}^\T$ and $\J \cdot \imtlg(\J) = \begin{bmatrix} -3&3&3 \end{bmatrix}^\T$.
$\imtlg$ is thus not non-conflicting.

It should be noted that when $\J$ has rank $m$, $\imtlg$ seems to be non-conflicting.
Thus, it would be possible to make a different non-conflicting generalization, for instance, by deciding $\0$ when $\J$ is not full rank.

\paragraph{\xmark\, Linear under scaling.}
With $\J = \begin{bmatrix}a & 0\\0 & 1\end{bmatrix}$ and $a > 0$, we have $\vv = \begin{bmatrix}1/a^2 & 0\\0 & 1\end{bmatrix} \cdot \begin{bmatrix}a \\ 1\end{bmatrix} = \begin{bmatrix}1/a \\ 1\end{bmatrix}$ and $\imtlg(\J) = \begin{bmatrix}a & 0\\0 & 1\end{bmatrix} \cdot \begin{bmatrix}1/a\\1\end{bmatrix} \cdot \frac{1}{\frac{1}{a} + 1} = \frac{1}{\frac{1}{a} + 1} \cdot \begin{bmatrix}1\\1\end{bmatrix}$.
This is not affine in $a$, so $\imtlg$ is not linear under scaling.

\paragraph{\cmark\, Weighted.}
By (\ref{eqn:imtlg}), $\imtlg$ is weighted.

\subsection{CAGrad}
$\cagrad$ is described in~\citet{cagrad}.
It is parameterized by $c\in\left[0,1\right[{}$.
If $c=0$, this is equivalent to $\meanagg$.
Therefore, we restrict our analysis to the case $c>0$.
For any $\J\in\Rmn$, let $\bar \vg$ be the average gradient $\frac{1}{m} \J^\T\cdot \1$, and let $\ve{i}^\T\J$ denote the $i$'th row of $\J$.
The aggregation is then defined as
\begin{equation}
\cagrad(\J) \in \argmax_{\substack{\vd\in\Rn:\\\|\vd-\overline \vg\|\leq c \|\overline \vg\|}}\min_{i\in \range{m}} \ve{i}^\T \J \vd
\end{equation}

\paragraph{\xmark\, Non-conflicting.}
Let $\J=\begin{bmatrix} 2&0\\-2a-2&2\end{bmatrix}$, with $a$ satisfying $-a+c\sqrt{a^2+1}<0$.
We have $\overline \vg=\begin{bmatrix} -a&1 \end{bmatrix}^\T$ and $\| \overline \vg\|=\sqrt{a^2+1}$.
Observe that any $\vd\in\Rn$ satisfying the constraint $\| \vd-\overline \vg\|\leq c \|\overline \vg\|$ has first coordinate at most $-a+c \sqrt{a^2+1}$.
Because $-a+c\sqrt{a^2+1}<0$, any feasible $d$ has a negative first coordinate, making $d$ conflict with the first row of $\J$.
For any $c \in [0, 1[$, $-a+c\sqrt{a^2+1}<0$ is equivalent to $\sqrt{\frac{c^2}{1-c^2}} < a$.
Thus, this provides a counter-example to the non-conflicting property for any $c \in [0, 1[$, \ie $\cagrad$ is not non-conflicting.

If we generalize to the case $c \geq 1$, as suggested in the original paper, then $\vd=\0$ becomes feasible, which yields $\min_{i\in \range{m}} \ve{i}^\T \J \vd = 0$.
Therefore the optimal $\vd$ satisfies $0\leq \min_{i\in \range{m}} \ve{i}^\T \J \vd$, \ie $\0\pleq \J \vd$.
With $c \geq 1$, $\cagrad$ would thus be non-conflicting.

\paragraph{\xmark\, Linear under scaling (sketch of proof).}
Let $\J=\begin{bmatrix} 2&0\\0&2a\end{bmatrix}$, then $\overline \vg=\begin{bmatrix} 1&a \end{bmatrix}^\T$ and $\| \overline \vg\|=\sqrt{1+a^2}$.
One can show that the constraint $\| \vd-\overline \vg\|\leq c \| \overline \vg\|$ needs to be satisfied with equality since, otherwise, we can scale $\vd$ to make the objective larger.
Substituting $\J$ in $\min_{i\in\range{m}} \ve{i}^\T \J \vd$ yields $2\min(d_1, ad_2)$.
For any $a$ satisfying $c \sqrt{1+a^2}+1 < a^2$, it can be shown that the optimal $\vd$ satisfies $d_1 < a d_2$.
In that case the inner minimum over $i$ is $2d_1$ and, to satisfy $\|\vd - \overline \vg\| = c \|\overline \vg\|$, the KKT conditions over the Lagrangian yield $\vd-\overline \vg ~\propto~ \nabla_\vd d_1=\begin{bmatrix} 1&0 \end{bmatrix}^\T$.
This yields $\vd=\begin{bmatrix}c \cdot\| \overline \vg\|+1\\a \end{bmatrix}=\begin{bmatrix}c \sqrt{1+a^2}+1\\a \end{bmatrix}$.
This is not affine in $a$; therefore, $\cagrad$ is not linear under scaling.

\paragraph{\cmark\, Weighted.}
In~\citet{cagrad}, $\cagrad$ is formulated via its dual: $\cagrad(\J) = \frac{1}{m} \J^\T \left(\1 + \frac{c \| \J^\T \1 \|}{\| \J^\T \vw\|} \vw \right)$, with $\vw\in \argmin_{\substack{\vw \in \Delta(m)}} \1^\T \J \J^\T \vw + c\cdot \| \J^\T \1 \|\cdot \| \J^\T \vw \|$, where $\Delta(m)$ is the probability simplex of dimension $m$.
Therefore, $\cagrad$ is weighted.

\subsection{RGW}
$\rgw$ is defined in~\citet{random} as the weighted sum of the rows of the input matrix, with a random weighting.
The weighting is obtained by sampling $m$ \iid normally distributed random variables and applying a softmax.
Formally, 
\begin{align}
\rgw(\J) &= \J^\T \cdot \softmax(\rvw)\label{eqn:rgw}\\
\text{with}\hspace{0.45cm}\rvw&\sim \mathcal N(\0,I)
\end{align}

\paragraph{\xmark\, Non-conflicting.}
When $\J = \begin{bmatrix} 1 \\ -2 \end{bmatrix}$, the only non-conflicting solution is $\0$. However, $\mathds{P}[\rgw(\J) = \0] = 0$, \ie $\rgw(\J)$ is in the dual cone of the rows of $\J$ with probability $0$.
$\rgw$ is thus not non-conflicting.
Here, 
\begin{align*}
\mathds{E}[\rgw(\J)] &= \mathbb E\left[\frac{e^{\ervw_1}-2e^{\ervw_2}}{e^{\ervw_1}+e^{\ervw_2}}\right]&\left(\substack{\text{By (\ref{eqn:rgw})}}\right)&\\
&=-\mathbb E\left[\frac{e^{\ervw_1}}{e^{\ervw_1}+e^{\ervw_2}}\right]&\left(\substack{\rvw\sim \mathcal N(\0,I)}\right)&\\
&<0&\left(\substack{0\plt e^{\rvw}}\right)&
\end{align*}
so $\rgw$ is neither non-conflicting in expectation.

\paragraph{\cmark\, Linear under scaling.}
$\rgw\big(\diag(\vc) \cdot \J\big) = \big(\diag(\vc) \cdot \J\big)^\T \cdot \softmax(\rvw) = \J^\T \cdot \diag(\vc) \cdot \softmax(\rvw) = \J^\T \cdot \diag\big(\softmax(\rvw)\big) \cdot \vc$.
We thus have $\rgw\big(\diag(\vc) \cdot \J\big) = \mathfrak{\J}^\T\cdot \vc$ with $\mathfrak{\J}^\T = \J^\T \cdot \diag\big(\softmax(\rvw)\big)$.
Therefore, $\rgw$ is linear under scaling.

\paragraph{\cmark\, Weighted.}
By (\ref{eqn:rgw}), $\rgw$ is weighted.

\subsection{Nash-MTL}
Nash-MTL is described in~\citet{nashmtl}.
Unfortunately, we were not able to verify the proof of Claim 3.1, and we believe that the official implementation of Nash-MTL may mismatch the desired objective by which it is defined.
Therefore, we only analyze the initial objective even though our experiments for this aggregator are conducted with the official implementation.

Let $\J\in\Rmn$ and $\varepsilon>0$. Let also $B_\varepsilon=\big\{ \vd\in\Rn : \| \vd \| \leq \varepsilon, \0\pleq \J \vd \big\}$. With $\ve{i}^\T\J$ denoting the $i$'th row of $\J$, $\nashmtl$ is then defined as
\begin{align}
\nashmtl(\J) = \argmax_{\vd\in B_\varepsilon} \sum_{i\in\range{m}} \log\big(\ve{i}^\T \J \vd\big)\label{eqn:nashmtl}
\end{align}

\paragraph{\cmark\, Non-conflicting.}
By the constraint, $\nashmtl$ is non-conflicting.

\paragraph{\xmark\, Linear under scaling.}
If an aggregator $\A$ is linear under scaling, it should be the case that $\A(a \J) = a \A(\J)$ for any scalar $a>0$ and any $\J\in\Rmn$.
However, $\log(a \ve{i}^\T \J \vd)=\log(\ve{i}^\T \J \vd) + \log(a)$. This means that scaling by a scalar does not impact aggregation.
Since this is not the trivial $\0$ aggregator, $\nashmtl$ is not linear under scaling.

\paragraph{\cmark\, Weighted.}
Suppose towards contradiction that $\vd$ is both optimal for (\ref{eqn:nashmtl}) and not in the span of $\J^\T$.
Let $\vd'$ be the projection of $\vd$ onto the span of  $\J^\T$.
Since $\| \vd' \| < \| \vd \|\leq\varepsilon$ and $\J \vd=\J \vd'$, we have $\J \vd\plt\J \left( \frac{\| \vd\|}{\|\vd'\|} \vd'\right)$, contradicting the optimality of $\vd$.
Therefore, $\nashmtl$ is weighted.

\subsection{Aligned-MTL}
The Aligned-MTL method for balancing the Jacobian is described in~\citet{aligned_mtl}.
For simplicity, we fix the vector of preferences to $\frac{1}{m} \1$, but the proofs can be adapted for any non-trivial vector.
Given $\J\in\Rmn$, let $V\Sigma^2 V^\T$ be the eigen-decomposition of $\J\J^\T$, let $\Sigma^\dagger$ be the diagonal matrix whose non-zero elements are the inverse of corresponding non-zero diagonal elements of $\Sigma$ and let $\sigma_{\min} = \min_{i\in\range{m},\Sigma_{ii}\neq 0} \Sigma_{ii}$.
The aggregation is then defined as
\begin{align}
\alignedmtl(\J) &= \frac{1}{m}\J^\T\cdot \vw\label{eqn:alignedmtl}\\
\text{with}\hspace{1.25cm}\vw&=\sigma_{\min}\cdot V\Sigma^\dagger V^\T\cdot \1
\end{align}

\paragraph{\xmark\, Non-conflicting.}
If the SVD of $\J$ is $V\Sigma U^\T$, then $\J^\T \vw =\sigma_{\min} U P V^\T \1$ with $P=\Sigma^\dagger \Sigma$ a diagonal projection matrix with $1$s corresponding to non zero elements of $\Sigma$ and $0$s everywhere else.
Further, $\J \cdot \alignedmtl(\J) = \frac{\sigma_{\min}}{m} \cdot V\Sigma V^\T \1$.
If $V=\frac{1}{2} \begin{bmatrix} \sqrt{3}&1\\-1&\sqrt{3} \end{bmatrix}$ and $\Sigma=\begin{bmatrix} 1&0\\0&0 \end{bmatrix}$, we have $\J \cdot \alignedmtl(\J)=\frac{1}{2} \cdot V\Sigma V^\T \1=\frac{1}{8}\begin{bmatrix} 3-\sqrt3&1-\sqrt3 \end{bmatrix}^\T$ which is not non-negative.
$\alignedmtl$ is thus not non-conflicting.

\paragraph{\xmark\, Linear under scaling.}
If $\J= \begin{bmatrix} 1&0\\0&a \end{bmatrix}$, then $U=V=I$, $\Sigma=\J$.
For $0<a\leq 1$, $\sigma_{\min}=a$, therefore $\alignedmtl(\J) = \frac{a}{2}\cdot \1$.
For $1<a$, $\sigma_{\min}=1$ and therefore $\alignedmtl(\J) = \frac{1}{2}\cdot\1$, which makes $\alignedmtl$ not linear under scaling.

\paragraph{\cmark\, Weighted.}
By (\ref{eqn:alignedmtl}), $\alignedmtl$ is weighted.
\clearpage
\section{Experimental settings}\label{appendix:experimental_settings}
For all of our experiments, we used \pytorch~\citep{pytorch}.
We have developed an open-source library\footnote{Available at \link{https://github.com/SimplexLab/TorchJD}} on top of it to enable Jacobian descent easily.
This library is designed to be reusable for many other use cases than the experiments presented in our work.
To separate them from the library, the experiments have been conducted with a different code repository\footnote{Available at \link{https://github.com/ValerianRey/jde}} mainly using \pytorch\ and our library.

\subsection{Learning rate selection}\label{appendix:lr_selection}
The learning rate has a very important impact on the speed of optimization.
To make the comparisons as fair as possible, we always show the results corresponding to the best learning rate.
Concretely, for each random rerun and each aggregator, we first try 22 learning rates from $10^{-5}$ to $10^2$, increasing by a factor $10^{\frac{1}{3}}$ every time.
The two best learning rates from this range then define a refined range of plausible good learning rates, going from the smallest of those two multiplied by $10^{-\frac{1}{3}}$ to the largest of those two multiplied by $10^{\frac{1}{3}}$.
This margin makes it unlikely for the best learning rate to lie out of the refined range.
After this, 50 learning rates from the refined range are tried.
These learning rates are evenly spaced in the exponent domain.
The best one is then selected and presented in the plots.
For simplicity, we have always used a constant learning rate, \ie no learning rate scheduler was used.

This approach has the advantage of being simple and precise, thus giving trustworthy results.
However, it requires 72 trainings for each aggregator, random rerun, and dataset, \ie a total of $43776$ trainings for all of our experiments.
For this reason, we have opted to work on small subsets of the original datasets.

\subsection{Random reruns and standard error of the mean}\label{appendix:random_reruns}
To get an idea of confidence in our results, every experiment is performed \nruns\ times on a different seed and a different subset, of size 1024, of the training dataset.
The seed used for run $i \in [\nruns]$ is always simply set to $i$.
Because each random rerun includes the full learning rate selection method described in Appendix~\ref{appendix:lr_selection}, it is sensible to consider the \nruns\ sets of results as \iid.
For each point of both the loss curves and the cosine similarity curves, we thus compute the estimated standard error of the mean with the usual formula $\frac{1}{\sqrt{\nruns}}\sqrt{\frac{\sum_{i\in\range{\nruns}}(v_i - \bar v)^2}{\nruns-1}}$, where $v_i$ is the value of a point of the curve for random rerun $i$, and $\bar v$ is the average value of this point over the \nruns\ runs.

\subsection{Model architectures}\label{appendix:model_architectures}
In all experiments, the models are simple convolutional neural networks.
All convolutions always have a stride of 1$\times$1, a kernel size of 3$\times$3, a learnable bias, and no padding.
All linear layers always have a learnable bias.
The activation function is the exponential linear unit~\citep{elu}.
The full architectures are given in Tables~\ref{tab:svhn_architecture},~\ref{tab:cifar10_architecture},~\ref{tab:eurosat_architecture} and~\ref{tab:mnist_architecture}.
Note that these architectures have been fixed arbitrarily, \ie they were not optimized through some hyper-parameter selection.
The weights of the model have been initialized with the default initialization scheme of \pytorch.

\begin{table}[p]
\caption{Architecture used for SVHN}
\label{tab:svhn_architecture}
\centering
\begin{center}
\begin{tabular}{l}
\toprule
\texttt{Conv2d} (3 input channels, 16 output channels, 1 group), \texttt{ELU}\\
\texttt{Conv2d} (16 input channels, 32 output channels, 16 groups)\\
\texttt{MaxPool2d} (stride of 2$\times$2, kernel size of 2$\times$2), \texttt{ELU}\\
\texttt{Conv2d} (32 input channels, 32 output channels, 32 groups)\\
\texttt{MaxPool2d} (stride of 3$\times$3, kernel size of 3$\times$3), \texttt{ELU}, \texttt{Flatten}\\
\texttt{Linear} (512 input features, 64 output features), \texttt{ELU}\\
\texttt{Linear} (64 input features, 10 outputs)\\
\bottomrule
\end{tabular}
\end{center}
\end{table}

\begin{table}[p]
\caption{Architecture used for CIFAR-10}
\label{tab:cifar10_architecture}
\centering
\begin{center}
\begin{tabular}{l}
\toprule
\texttt{Conv2d} (3 input channels, 32 output channels, 1 group), \texttt{ELU}\\
\texttt{Conv2d} (32 input channels, 64 output channels, 32 groups)\\
\texttt{MaxPool2d} (stride of 2$\times$2, kernel size of 2$\times$ 2), \texttt{ELU}\\
\texttt{Conv2d} (64 input channels, 64 output channels, 64 groups)\\
\texttt{MaxPool2d} (stride of 3$\times$3, kernel size of 3$\times$3), \texttt{ELU}, \texttt{Flatten}\\
\texttt{Linear} (1024 input features, 128 output features), \texttt{ELU}\\
\texttt{Linear} (128 input features, 10 outputs)\\
\bottomrule
\end{tabular}
\end{center}
\end{table}

\begin{table}[p]
\caption{Architecture used for EuroSAT}
\label{tab:eurosat_architecture}
\centering
\begin{center}
\begin{tabular}{l}
\toprule
\texttt{Conv2d} (3 input channels, 32 output channels, 1 group)\\
\texttt{MaxPool2d} (stride of 2$\times$2, kernel size of 2$\times$2), \texttt{ELU}\\
\texttt{Conv2d} (32 input channels, 64 output channels, 32 groups)\\
\texttt{MaxPool2d} (stride of 2$\times$2, kernel size of 2$\times$2), \texttt{ELU}\\
\texttt{Conv2d} (64 input channels, 64 output channels, 64 groups)\\
\texttt{MaxPool2d} (stride of 3$\times$3, kernel size of 3$\times$3), \texttt{ELU}, \texttt{Flatten}\\
\texttt{Linear} (1024 input features, 128 output features), \texttt{ELU}\\
\texttt{Linear} (128 input features, 10 outputs)\\
\bottomrule
\end{tabular}
\end{center}
\end{table}

\begin{table}[p]
\caption{Architecture used for MNIST, Fashion-MNIST and Kuzushiji-MNIST}
\label{tab:mnist_architecture}
\centering
\begin{center}
\begin{tabular}{l}
\toprule
\texttt{Conv2d} (1 input channel, 32 output channels, 1 group), \texttt{ELU}\\
\texttt{Conv2d} (32 input channels, 64 output channels, 1 group)\\
\texttt{MaxPool2d} (stride of 2$\times$2, kernel size of 2$\times$2), \texttt{ELU}\\
\texttt{Conv2d} (64 input channels, 64 output channels, 1 group)\\
\texttt{MaxPool2d} (stride of 3$\times$3, kernel size of 3$\times$3), \texttt{ELU}, \texttt{Flatten}\\
\texttt{Linear} (576 input features, 128 output features), \texttt{ELU}\\
\texttt{Linear} (128 input features, 10 outputs)\\
\bottomrule
\end{tabular}
\end{center}
\end{table}

\subsection{Optimizer}
For all experiments except those described in Appendix~\ref{appendix:adam_study}, we always use the basic \texttt{SGD} optimizer of \pytorch, without any regularization or momentum.
Here, \texttt{SGD} refers to the \pytorch\ optimizer that updates the parameters of the model in the opposite direction of the gradient, which, in our case, is replaced by the aggregation of the Jacobian matrix.
In the rest of this paper, SGD refers to the whole stochastic gradient descent algorithm.
In the experiments of Appendix~\ref{appendix:adam_study}, we instead use \texttt{Adam} to study its interactions with JD.

\subsection{Loss function}
The loss function is always the usual cross-entropy, with the default parameters of \pytorch.

\subsection{Preprocessing}

The inputs are always normalized per channel based on the mean and standard deviation computed on the entire training split of the dataset.

\subsection{Iterations and computational budget}\label{appendix:budget}

The numbers of epochs and the corresponding numbers of iterations for all datasets are provided in Table~\ref{tab:epochs}, along with the required number of NVIDIA L4 GPU-hours, to run all 72 learning rates for the 11 aggregators on a single seed.
The total computational budget to run the main experiments on \nruns\ seeds was thus around 760 GPU-hours.
Additionally, we used a total of about 100 GPU hours for the experiments varying the batch size and using Adam, and about 200 more GPU hours were used for early investigations.

\begin{table}[ht!]
\caption{Numbers of epochs, iterations, and GPU-hours for each dataset}
\label{tab:epochs}
\centering
\begin{center}
\begin{tabular}{lccc}
\toprule
Dataset & Epochs & Iterations & GPU-Hours\\
\midrule
SVHN & 25 & 800 & 17\\
CIFAR-10 & 20 & 640 & 15\\
EuroSAT & 30 & 960 & 32\\
MNIST & 8 & 256 & 6\\
Fashion-MNIST & 25 & 800 & 17\\
Kuzushiji-MNIST & 10 & 320 & 8\\
\bottomrule
\end{tabular}
\end{center}
\end{table}

\clearpage
\section{Computation time and memory usage}\label{appendix:speed}

\subsection{Empirical memory usage}
Table~\ref{tab:memory} compares the practical memory usage of the different methods.
Since we used the same architecture for MNIST, Fashion-MNIST and Kuzushiji-MNIST, we only report the results for one of them.
With the Gramian-based method proposed in Section~\ref{sec:implementation}, only the Gramian, which is typically small, has to be stored: the memory requirement becomes much closer to that of SGD.
We believe that a better implementation of the Gramian-based approach can reduce this memory cost even further, to the level of SGD's memory usage.

\begin{table}[ht!]
\rowcolors{2}{table_odd_row_color}{}
\caption{Peak GPU memory used during one epoch of training on the ERM objective with SGD, on the IWRM objective with SSJD, and on the IWRM objective with Gramian-based SSJD (as described in Section~\ref{sec:implementation}). The batch size is always 32.}
\label{tab:memory}
\centering
\begin{center}
\begin{tabular}{llcccc}
\toprule
Objective & Method & SVHN & CIFAR-10 & EuroSAT & MNIST\\
\midrule
ERM  & SGD                      & 32.0 MB  & 44.6 MB  & 72.6 MB   & 54.2 MB  \\
IWRM & SSJD                     & 315.4 MB & 619.3 MB & 1447.5 MB & 786.0 MB \\
IWRM & SSJD (Gramian-based)     & 39.8 MB  & 75.3 MB  & 104.2 MB  & 65.6 MB  \\
\bottomrule
\end{tabular}
\end{center}
\end{table}

\subsection{Empirical computational times}
In Table~\ref{tab:speed}, we compare the computation time of SGD with that of SSJD for all the aggregators that we experimented with.
Since we used the same architecture for MNIST, Fashion-MNIST and Kuzushiji-MNIST, we only report the results for one of them.
Several factors affect this computation time.
First, the batch size affects the number of rows in the Jacobian to aggregate.
Increasing the batch size thus requires more GPU memory and the aggregation of a taller matrix.
In contrast, when using the Gramian-based implementation, the memory usage is greatly reduced, which, on a GPU, improves parallelization and thus computational time.
Then, some aggregators, \eg $\nashmtl$ and $\mgda$, seem to greatly increase the run time.
When the aggregation is the bottleneck, a faster implementation will be necessary to make them usable in practice.
Lastly, our current implementation of Gramian-based JD and of $\upgrad$ can still be greatly improved.

\begin{table}[ht!]
\rowcolors{2}{table_odd_row_color}{}
\caption{Time required in seconds for one epoch of training on the ERM objective with SGD, on the IWRM objective with SSJD, and on the IWRM objective with Gramian-based SSJD (as described in Section~\ref{sec:implementation}), for different aggregators, on an NVIDIA L4 GPU. The batch size is always 32.}
\label{tab:speed}
\centering
\begin{center}
\begin{tabular}{llcccc}
\toprule
Objective & Method & SVHN & CIFAR-10 & EuroSAT & MNIST\\
\midrule
ERM  & SGD                               & 0.83 & 0.44 & 0.76 & 0.43 \\
IWRM & SSJD--$\upgrad$                   & 1.63 & 1.78 & 2.86 & 1.74 \\
IWRM & SSJD--$\upgrad$ (Gramian-based)     & 1.15 & 1.12 & 1.51 & 1.11 \\

IWRM & SSJD--$\meanagg$                  & 1.36 & 1.67 & 2.71 & 1.62 \\
IWRM & SSJD--$\meanagg$ (Gramian-based)  & 1.01 & 1.00 & 1.32 & 0.97 \\

IWRM & SSJD--$\mgda$                     & 3.92 & 3.95 & 5.36 & 4.05 \\
IWRM & SSJD--$\mgda$ (Gramian-based)       & 3.56 & 3.31 & 3.98 & 3.41 \\

IWRM & SSJD--$\dualproj$                 & 1.40 & 1.69 & 2.72 & 1.64 \\
IWRM & SSJD--$\dualproj$ (Gramian-based)   & 1.03 & 1.02 & 1.33 & 1.00 \\

IWRM & SSJD--$\pcgrad$                   & 2.55 & 2.89 & 3.89 & 2.83 \\
IWRM & SSJD--$\pcgrad$ (Gramian-based)     & 2.20 & 2.22 & 2.50 & 2.20 \\

IWRM & SSJD--$\graddrop$                 & 1.34 & 1.69 & 2.67 & 1.59 \\

IWRM & SSJD--$\imtlg$                    & 1.43 & 1.69 & 2.73 & 1.65 \\
IWRM & SSJD--$\imtlg$ (Gramian-based)      & 1.04 & 1.03 & 1.33 & 1.00 \\

IWRM & SSJD--$\cagrad$                   & 1.74 & 2.02 & 3.08 & 1.98 \\
IWRM & SSJD--$\cagrad$ (Gramian-based)     & 1.36 & 1.35 & 1.68 & 1.34 \\

IWRM & SSJD--$\rgw$                      & 1.38 & 1.67 & 2.70 & 1.62 \\
IWRM & SSJD--$\rgw$ (Gramian-based)        & 1.00 & 1.00 & 1.30 & 0.97 \\

IWRM & SSJD--$\nashmtl$                  & 5.72 & 6.20 & 7.06 & 6.13 \\

IWRM & SSJD--$\alignedmtl$               & 1.49 & 1.73 & 2.75 & 1.68 \\
IWRM & SSJD--$\alignedmtl$ (Gramian-based) & 1.06 & 1.05 & 1.36 & 1.03 \\

\bottomrule
\end{tabular}
\end{center}
\end{table}

\subsection{Time complexity of the unconflicting projection of gradients}
Let $\J \in \Rmn$ be the matrix to aggregate.
Apart from computing the Gramian $\J\J^\T$, applying $\upgrad$ to $\J$ requires solving $m$ instances of the quadratic program (\ref{eqn:upgrad_dual}) of Proposition~\ref{prop:upgrad_duality}.
Solvers for such problems of dimension $m$ typically have a computational complexity upper bounded by $\bigo(m^4)$ or less in recent implementations (\eg $\bigo(m^{3.67}\log m)$).
This induces a $\bigo(m^5)$ time complexity for extracting the weights of $\upgrad$.
Note that solving these $m$ problems in parallel would reduce this complexity to $\bigo(m^4)$.

\clearpage
\section*{NeurIPS Paper Checklist}

\begin{enumerate}

\item {\bf Claims}
    \item[] Question: Do the main claims made in the abstract and introduction accurately reflect the paper's contributions and scope?
    \item[] Answer: \answerYes{} 
    \item[] Justification: We have stated our contributions as accurately as we could in the abstract and in the introduction. In particular, we have tried to clarify that Jacobian descent is more of a formalized unification of existing methods, rather than an entirely new concept.

\item {\bf Limitations}
    \item[] Question: Does the paper discuss the limitations of the work performed by the authors?
    \item[] Answer: \answerYes{} 
    \item[] Justification: The main limitation of Jacobian descent is the memory requirement, and to some extent, the computational cost. We discuss this issue at the end of Section~\ref{sec:experiments}. Section~\ref{sec:implementation} provides promising future directions to address this. The limited empirical evaluation is also discussed in a dedicated paragraph at the end of the conclusion.

\item {\bf Theory assumptions and proofs}
    \item[] Question: For each theoretical result, does the paper provide the full set of assumptions and a complete (and correct) proof?
    \item[] Answer: \answerYes{} 
    \item[] Justification: All proofs are available in the appendices and were thoroughly reviewed. In Appendix \ref{appendix:supplementary_theory}, we provide additional results for reference. In all of Appendix \ref{appendix:proofs}, we give proofs of the stated results, together with Lemmas that help proving them. In Appendix \ref{appendix:properties}, we provide justifications for the results stated in Table \ref{tab:properties}, about properties of many existing aggregators.

    \item {\bf Experimental result reproducibility}
    \item[] Question: Does the paper fully disclose all the information needed to reproduce the main experimental results of the paper to the extent that it affects the main claims and/or conclusions of the paper (regardless of whether the code and data are provided or not)?
    \item[] Answer: \answerYes{} 
    \item[] Justification: We have provided a very detailed explanation of the experimental settings in Appendix~\ref{appendix:experimental_settings}. Besides, the code necessary to run our experiments is publicly available at \link{https://github.com/ValerianRey/jde}. To make Jacobian descent even more reproducible, we have published a PyTorch library for it, available at \link{https://github.com/SimplexLab/TorchJD}.

\item {\bf Open access to data and code}
    \item[] Question: Does the paper provide open access to the data and code, with sufficient instructions to faithfully reproduce the main experimental results, as described in supplemental material?
    \item[] Answer: \answerYes{} 
    \item[] Justification: The code used for our experiments is publicly available at \link{https://github.com/ValerianRey/jde}. It can be used to reproduce our results, by following the instructions of the README file. The data always comes from torchvision, so it will be downloaded automatically as needed.

\item {\bf Experimental setting/details}
    \item[] Question: Does the paper specify all the training and test details (e.g., data splits, hyperparameters, how they were chosen, type of optimizer) necessary to understand the results?
    \item[] Answer: \answerYes{} 
    \item[] Justification: The experimental settings are all given in Appendix~\ref{appendix:experimental_settings}. We have tried to give as many details as possible, and to justify our choices.

\item {\bf Experiment statistical significance}
    \item[] Question: Does the paper report error bars suitably and correctly defined or other appropriate information about the statistical significance of the experiments?
    \item[] Answer: \answerYes{} 
    \item[] Justification: In our experiments, we show the standard error of the mean to get some idea of confidence in our results. We explicitly state how it is computed in Appendix~\ref{appendix:random_reruns}. We also provide box plots of the test set accuracies in Appendix~\ref{appendix:test_results}.

\item {\bf Experiments compute resources}
    \item[] Question: For each experiment, does the paper provide sufficient information on the computer resources (type of compute workers, memory, time of execution) needed to reproduce the experiments?
    \item[] Answer: \answerYes{} 
    \item[] Justification: We have included in Appendix~\ref{appendix:budget} the computational resources that we have used.
    
\item {\bf Code of ethics}
    \item[] Question: Does the research conducted in the paper conform, in every respect, with the NeurIPS Code of Ethics \url{https://neurips.cc/public/EthicsGuidelines}?
    \item[] Answer: \answerYes{} 
    \item[] Justification: Our paper should not cause any predictable harm or have potential harmful consequences. We have spent substantial efforts in making it beneficial to the community. For instance, we provide a library to experiment with Jacobian descent that we documented and tested extensively.

\item {\bf Broader impacts}
    \item[] Question: Does the paper discuss both potential positive societal impacts and negative societal impacts of the work performed?
    \item[] Answer: \answerNA{} 
    \item[] Justification: Our work is mostly theoretical, so we do not think that it has direct societal impacts.
    
\item {\bf Safeguards}
    \item[] Question: Does the paper describe safeguards that have been put in place for responsible release of data or models that have a high risk for misuse (e.g., pre-trained language models, image generators, or scraped datasets)?
    \item[] Answer: \answerNA{} 
    \item[] Justification: To the best of our knowledge, our work presents no such risks.

\item {\bf Licenses for existing assets}
    \item[] Question: Are the creators or original owners of assets (e.g., code, data, models), used in the paper, properly credited and are the license and terms of use explicitly mentioned and properly respected?
    \item[] Answer: \answerYes{} 
    \item[] Justification: We have always credited in our code the original implementation from which code was taken and adapted, when applicable. We have cited the corresponding papers in our main text, and we have respected the terms of their licenses. In particular, for MIT licenses, we have duly copied the license at the top of the concerned files.

\item {\bf New assets}
    \item[] Question: Are new assets introduced in the paper well documented and is the documentation provided alongside the assets?
    \item[] Answer: \answerYes{} 
    \item[] Justification: We spent a substantial amount of time to provide a high quality documentation of our library. We hope that it will help the community to experiment with the concepts that we introduced in our work.

\item {\bf Crowdsourcing and research with human subjects}
    \item[] Question: For crowdsourcing experiments and research with human subjects, does the paper include the full text of instructions given to participants and screenshots, if applicable, as well as details about compensation (if any)? 
    \item[] Answer: \answerNA{} 
    \item[] Justification: Our work did not involve any crowdsourcing nor human subject.

\item {\bf Institutional review board (IRB) approvals or equivalent for research with human subjects}
    \item[] Question: Does the paper describe potential risks incurred by study participants, whether such risks were disclosed to the subjects, and whether Institutional Review Board (IRB) approvals (or an equivalent approval/review based on the requirements of your country or institution) were obtained?
    \item[] Answer: \answerNA{} 
    \item[] Justification: Our work did not involve any human subject.

\item {\bf Declaration of LLM usage}
    \item[] Question: Does the paper describe the usage of LLMs if it is an important, original, or non-standard component of the core methods in this research? Note that if the LLM is used only for writing, editing, or formatting purposes and does \emph{not} impact the core methodology, scientific rigor, or originality of the research, declaration is not required.
    \item[] Answer: \answerNA{} 
    \item[] Justification: We only used LLMs for some minor reformulations when writing, and to help phrase some of the revisions made in response to the reviewers' suggestions.

\end{enumerate}

\end{document}